\documentclass[10pt]{article}
\usepackage[letterpaper,margin=1in]{geometry}
\usepackage[T1]{fontenc}
\usepackage[utf8]{inputenc}
\usepackage{times}
\usepackage{amsmath,amssymb,amsfonts,bm}
\usepackage{graphicx}
\usepackage{booktabs,multirow,array,tabularx}
\usepackage[table]{xcolor}
\usepackage{subcaption}
\usepackage{enumitem}
\usepackage{microtype}
\usepackage{xspace}
\usepackage[numbers,sort&compress]{natbib}
\usepackage[colorlinks=true,linkcolor=blue!50!black,citecolor=blue!50!black,urlcolor=blue!50!black]{hyperref}
\usepackage[capitalize,noabbrev]{cleveref}
\usepackage{caption}
\setlist{itemsep=1pt,topsep=2pt,leftmargin=1.2em}
\definecolor{ours}{HTML}{E8F1FC}

\newcommand{\method}{\textsc{TASK}\xspace}
\newcommand{\T}{\textsf{T}\xspace}
\newcommand{\E}{\textsf{E}\xspace}
\newcommand{\Sv}{\textsf{S}\xspace}
\newcommand{\vp}{v^{+}}
\newcommand{\vm}{v^{-}}
\newcommand{\pp}{\,pp\xspace}
\newcommand{\ci}[2]{{\scriptsize[#1,\,#2]}}
\newcommand{\sig}{$^{*}$}

\title{\textbf{TASK Memory: Beyond \underline{T}ext-Only Memory with \underline{A}ctivation \underline{S}teering and \underline{K}nowledge Editing against Memory Hallucination}}

\newcommand{\AuthorOne}{Jinkun Zhao}
\newcommand{\AuthorTwo}{Wenjun Wu}
\newcommand{\AuthorThree}{Yanjun Pu}
\newcommand{\AuthorFour}{Junle Wang}
\newcommand{\AffilOne}{SKLCCSE, Institute of Artificial Intelligence Beihang University, Beijing, China}
\newcommand{\AffilTwo}{Beijing Advanced Innovation Center for Future Blockchain and Privacy Computing Beihang University, Beijing, China}
\newcommand{\EmailOne}{wwj09315@buaa.edu.cn}
\author{%
  \AuthorOne$^{1}$\quad \AuthorThree$^{1}$\quad \AuthorFour$^{1}$\quad \AuthorTwo$^{1,2}$ \quad
  \\[4pt]
  {\normalsize $^{1}$\AffilOne}\\
  {\normalsize $^{2}$\AffilTwo}\\[2pt]
  {\normalsize\texttt{\EmailOne}}%
}

\date{}

\begin{document}
\maketitle

\begin{abstract}
LLM agents increasingly rely on external memory, yet they still hallucinate about what they remember. Much of this is not a storage error: the correct memory is retrieved and placed in the prompt, but the answer contradicts it, or the memory is applied to the wrong question. We call these \emph{use-time memory hallucinations}. Across five open models, we find that they come from the conflict between the model's parametric knowledge and its external memory, and that they take three forms. In \emph{memory-ignoring} hallucination, the model returns to a stale fact stored in its weights despite the memory, in up to 46\% of queries. In \emph{memory-misapplying} hallucination, it attributes the memory to unrelated neighbour entities, in up to 24\% of cases, and in up to 49\% once retrieval is combined with weight editing. In \emph{version confusion}, 53\% of answers about revised facts assert the superseded value when both versions remain stored.
We propose \method memory, a memory that is not text alone: each memory combines a \textbf{T}ext note with \textbf{A}ctivation \textbf{S}teering and \textbf{K}nowledge editing, which act inside the model. The three carriers share one in-model address. The text \T holds the content. A local, removable knowledge edit \E supplies the address through its subject key and supports multi-hop composition through its weight change. A pair of activation-steering vectors \Sv decides, query by query, whether the model should adopt the memory or keep its own answer. \Sv stores no content; it controls how the model uses the text memory, rather than writing new memory into the parameters.
On frozen off-the-shelf models, under a pre-registered, test-once protocol, \method reduces memory-ignoring hallucinations to $\le$1.5\% in all ten model--dataset cells, and cuts neighbour misuse from 39.2\% to 7.5\% (+10.8\pp composite on CounterFact). It also improves a 300-memory library over BM25 RAG by 16.4\pp, an advantage that holds at 3{,}000 memories, resolves fact revisions without fact IDs, and adds 7.5\pp of multi-hop accuracy on MQuAKE. 21 of 23 cross-model comparisons (8B--27B, four families) are significant, and \method beats three recent context-steering methods in 9 of 10 model--dataset cells.
\end{abstract}

\section{Introduction}
\label{sec:intro}

\begin{figure}[t]
\centering
\includegraphics[width=\linewidth]{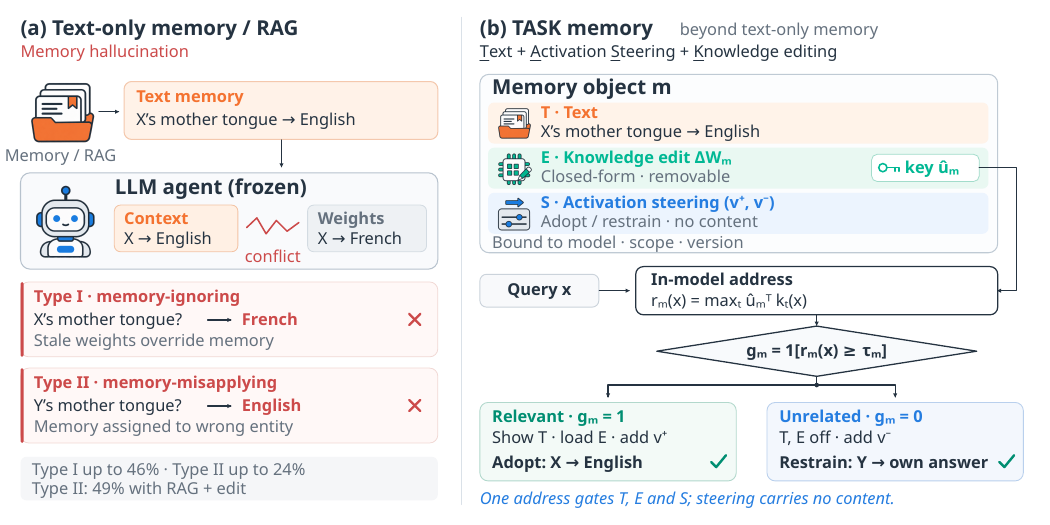}
\caption{\textbf{Memory hallucination: the memory is correct, the answer is not.} \textbf{(a)} Text-only memory or RAG puts the retrieved note into the prompt and leaves the rest to the frozen model, where the note in the context conflicts with the stale fact in the weights. On a question about the memory, the weights win and the model answers with the stale fact (Type I, memory-ignoring). On a question about another entity, the note wins and the model assigns the memory to the wrong entity (Type II, memory-misapplying). Across our five models these rates reach 46\% and 24\%, and Type II reaches 49\% when retrieval is combined with weight editing. \textbf{(b)} \method memory goes beyond text-only memory. Each memory object holds \textbf{T}ext content, a \textbf{K}nowledge edit $\Delta W_m$ whose key $\hat u_m$ serves as the in-model address, and a pair of \textbf{A}ctivation-\textbf{S}teering vectors $(\vp,\vm)$ that carry no content. For a query $x$, the address response $r_m(x)$ is compared with the threshold $\tau_m$. A relevant query shows \T, loads \E and adds $\vp$, so the model adopts the memory; an unrelated query leaves \T and \E off and adds $\vm$, so the model keeps its own answer.}
\label{fig:teaser}
\end{figure}

Agents built on large language models (LLMs) keep meeting facts that their weights do not contain, or contain in an outdated form: a person changes jobs, a user replaces a device, an API is deprecated. The dominant way to remember such facts is \emph{text memory}. The fact is written as a note, stored in a memory store, and retrieved into the prompt when a related question arrives~\citep{lewis2020rag,packer2023memgpt,park2023generative,zhong2024memorybank,chhikara2025mem0,xu2025amem}. Agents with memory still hallucinate. HaluMem~\citep{chen2025halumem} shows that memory systems introduce hallucinations when they extract and update memories, and that these errors carry into question answering. That diagnosis locates the error in what is \emph{stored}. Here we study the complementary failure: \emph{the stored memory is correct and is even in the prompt, yet the answer contradicts it, or the memory is applied where it does not belong.} We call these \textbf{use-time memory hallucinations}. They are a form of context-conflicting (unfaithful) hallucination~\citep{zhang2023siren,huang2025survey} in which the conflicting context is the agent's own memory.

\begin{figure}[t]
\centering
\includegraphics[width=\linewidth]{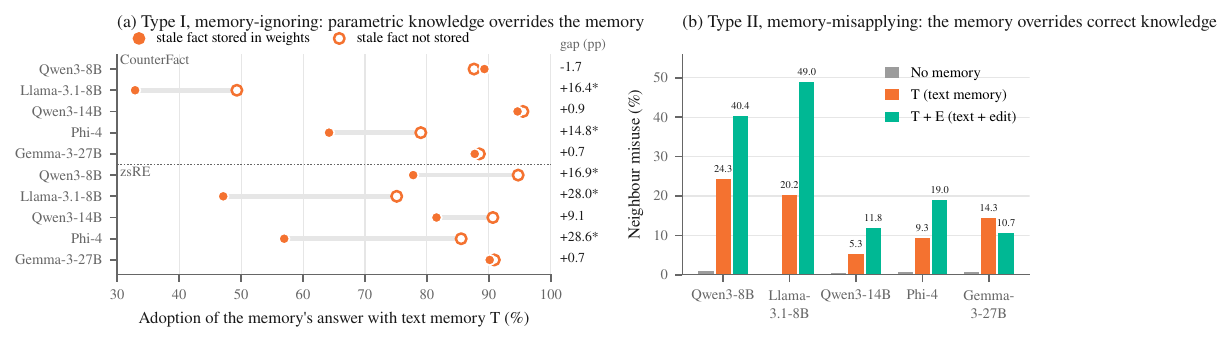}
\caption{\textbf{Use-time memory hallucinations caused by the conflict between parametric knowledge and external memory} (test splits, text memory in context, greedy decoding).
\textbf{(a)} Type I, memory-ignoring: adoption of the memory's answer on paraphrased questions. Filled markers: queries on which the no-memory model states the stale fact, i.e.\ the old fact is in the weights. Hollow markers: all other queries. The gap is in percentage points; $^{*}$ marks a paired world-level bootstrap 95\% CI that excludes 0.
\textbf{(b)} Type II, memory-misapplying: on neighbour questions (same relation, different subject) that the no-memory model answers correctly, the rate at which the model instead states the memory's answer. Without memory this is $\le$0.9\%. With text memory it rises on every model, and with text memory plus a weight edit (T+E) it rises further on four of the five models. All numbers come from frozen predictions (\cref{app:conflict}).}
\label{fig:conflict}
\end{figure}

Use-time memory hallucinations arise because the memory is not read by a blank slate. The model already holds parametric knowledge about the same entities, and the note and the weights compete inside every forward pass. Such knowledge conflicts are known from question answering over retrieved passages~\citep{longpre2021entity,xie2024chameleon,xu2024conflicts}. We measure them in the agent-memory setting, on five open models (Qwen3-8B/14B, Llama-3.1-8B, Phi-4, Gemma-3-27B) and three memory settings, and find three types (\cref{fig:conflict}). The first, \textbf{memory-ignoring hallucination} (Type I), is parametric override. When we split queries by whether the no-memory model still states the pre-update fact, text memory turns out to be adopted 14.8--28.6\pp less often on queries whose stale fact is stored in the weights, in five of ten model--dataset cells (\cref{fig:conflict}a). On those queries the model asserts the stale answer in 4.9--46.3\% of cases, even though the correct note is in the prompt. The second, \textbf{memory-misapplying hallucination} (Type II), runs in the opposite direction. On neighbour questions that share the relation but not the subject, and that the no-memory model answers correctly, the model attributes the note's answer to the wrong entity in 5.3--24.3\% of cases on CounterFact, against less than 1\% without memory. A natural ``RAG + editing'' design, which adds a weight edit to the text memory, raises this rate to as much as 49\% (\cref{fig:conflict}b). The third, \textbf{version-confusion hallucination} (Type III), appears once facts are revised: when both versions remain in the store, 53\% of answers about revised facts assert the superseded value (Qwen3-8B, BM25 RAG; \cref{app:hard}).

Because Types I and II pull in opposite directions, no single global remedy fixes both, and each existing family of methods addresses at most one of them. Better retrieval decides which note enters the prompt, but not how the model weighs it against its weights. Context-faithfulness methods, such as contrastive decoding~\citep{shi2024cad} and global steering directions~\citep{anand2026contextfocus,lou2026ksd,zhao2025spare}, push the model towards the context uniformly and thereby trade Type I for Type II; in our experiments an ungated evidence direction raises misuse and lowers the composite score below that of text memory alone (0.780 vs.\ 0.811). Knowledge editing~\citep{meng2022rome,meng2023memit,fang2025alphaedit} writes the fact into the weights, which amplifies Type II when it is combined with text memory (\cref{fig:conflict}b), and a joint closed-form patch over accumulating versions collapses (\cref{sec:rq2}). Memory-native models such as MemT-1.0~\citep{memtensor2026memt} go further and move the memory \emph{content} itself into the model: MemT-1.0 trains native memory blocks on 34B tokens, writes the interaction history into a dynamic memory state, and reads it back through a parallel memory attention. This changes \emph{where} the memory lives, but not \emph{who arbitrates} when the stored item and the weights disagree. Its memory modules are trained to imitate the backbone given the full textual history, so they inherit the backbone's way of using context, and in their own evaluation conflicts with real-world knowledge are handled by an instruction to answer ``from the knowledge pool \ldots\ rather than the real facts in real world''.

\begin{figure}[t]
\centering
\includegraphics[width=\linewidth]{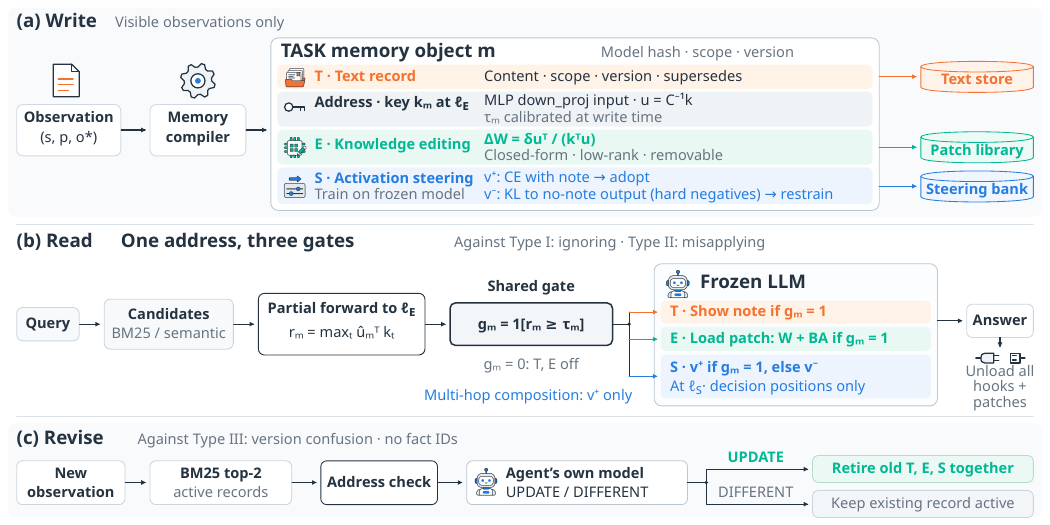}
\caption{\textbf{\method memory overview.} \textbf{(a) Write.} A visible observation $(s,p,o^*)$ is compiled into a memory object bound to the model hash, scope and version. The object contains a text record \T, an address key $k_m$ at layer $\ell_E$ with a threshold $\tau_m$ calibrated at write time, a closed-form, low-rank and removable patch \E keyed on the same address, and steering vectors \Sv\,$=(\vp_m,\vm_m)$ trained on the frozen model; these are kept in a text store, a patch library and a steering bank. \textbf{(b) Read}, against memory-ignoring (Type I) and memory-misapplying (Type II) hallucination. Candidates are retrieved by BM25 or semantic search, and one partial forward pass to layer $\ell_E$ computes each candidate's address response $r_m$. The shared gate $g_m=\mathbb 1[r_m\ge\tau_m]$ decides whether the note is shown, whether the patch is loaded as $W+BA$, and whether $\vp_m$ or $\vm_m$ is added at the decision positions of layer $\ell_S$; multi-hop composition uses $\vp$ only. All hooks and patches are unloaded after generation. \textbf{(c) Revise}, against version confusion (Type III). A new observation is matched to the top-2 active records by BM25, checked with the address and judged by the agent's own model. On \textsc{update} the old version's \T, \E and \Sv are retired together; on \textsc{different} the existing record stays active. No fact IDs are used.}
\label{fig:method}
\end{figure}

We take a different view: \emph{using} a memory is a separate, controllable decision, and use-time memory hallucination is a failure of that decision. For each query the model must decide whether a note should override what it already believes. That decision is made inside the network, so it can be read out and steered there. We therefore propose \method memory, which goes beyond text-only memory by coupling each \textbf{T}ext note with \textbf{A}ctivation \textbf{S}teering and \textbf{K}nowledge editing, two carriers that act inside the model and share one in-model address (\cref{fig:teaser,fig:method}). The text \T carries the content, a versioned note that tells the model \emph{what} the fact is. The activation-steering carrier \Sv carries the use policy: a pair of residual-stream vectors, of which $\vp$ makes the model adopt the note when the address fires, countering Type I, and $\vm$ makes it keep its parametric answer when the note is present but the query concerns something else, countering Type II. The knowledge-editing carrier \E is a closed-form, removable MLP edit keyed on the subject's internal representation~\citep{meng2022rome,geva2021ffn}. Its key doubles as the memory's address, so that one partial forward pass tells whether a query is about this memory, and its weight change supports multi-hop composition. At read time the address gates all three carriers at once, deciding which notes are shown, which patches are loaded and which steering direction is applied. Relevance is thus judged in one place, the model's own representation, rather than separately by a retriever, an editor and a steering rule. When a fact is revised, the old version's text, patch and vectors are retired together, which removes Type III.

The steering carrier is deliberately not a memory store. Without the text note it answers 1.7\% of CounterFact queries, the same as the unmodified model. What it changes is the arbitration: memory-ignoring hallucinations fall to at most 1.5\% in all ten model--dataset cells, and neighbour misuse falls from 39.2\% to 7.5\% (Qwen3-8B, CounterFact). This is the key difference from parametric-memory approaches such as knowledge editing, parametric RAG~\citep{tan2025dyprag} and MemT-1.0, which put new \emph{content} into weights or a trained memory state. \method keeps content in text, where it can be inspected, versioned and deleted, and writes into the model only a bidirectional controller that applies to a single memory and can be removed.

We evaluate \method memory on five frozen models from four families (8B--27B) and six benchmarks under a pre-registered protocol. On single facts it improves over text memory by 10.8\pp on CounterFact and 5.7\pp on zsRE (Qwen3-8B, composite of adoption and locality). For an agent holding 300 memories it improves over BM25 RAG by 16.4\pp and 16.2\pp, mainly by removing Type II hallucinations caused by retrieved distractors, and the advantage holds at 3{,}000 memories (0.926 for address-verified \T+\Sv vs.\ 0.731). Address-assisted supersession, which needs no fact IDs, raises accuracy on revised facts by 48.7--54.7\pp, and on MQuAKE the weight carrier adds 7.5\pp of multi-hop accuracy. Across models, 21 of 23 comparisons are significant, and \method beats knowledge-selection steering~\citep{lou2026ksd}, ContextFocus~\citep{anand2026contextfocus} and CAST~\citep{lee2025cast} in 9 of 10 model--dataset cells. The experiments also reveal a clear division of labour among the carriers: the weight change of \E matters for composition but not for single facts, and training the steering vectors on the edited model brings no reliable gain over training them on the base model.

\paragraph{Contributions.} First, we define use-time memory hallucination and its three types, and show on five models that memory-ignoring and memory-misapplying hallucinations arise from the two-way conflict between parametric knowledge and external memory, with weight edits added to retrieval amplifying the latter (\cref{fig:conflict}). Second, we propose \method memory, which combines text, activation steering and knowledge editing in one memory object: text carries the content, a removable edit provides an in-model address and compositional recall, and a paired adopt/restrain steering policy arbitrates each query, with the address gating prompt, weights and activations together across a write, read and revision lifecycle that detects revisions without fact IDs. Third, we provide pre-registered evidence on five models against twelve published baselines, together with an analysis showing that \Sv learns a shared adopt/restrain direction plus a memory-specific residual that is effective only behind the address gate.

\section{Related Work}
\label{sec:related}

\paragraph{Hallucination and memory hallucination.}
Hallucination surveys separate output that conflicts with world facts from output that conflicts with the input or context the model was given~\citep{ji2023survey,zhang2023siren,huang2025survey}. Agent memory adds a new source of context, so it adds new ways to hallucinate. HaluMem~\citep{chen2025halumem} evaluates memory systems operation by operation (extraction, updating, question answering). It finds that hallucinations are introduced and accumulate when memories are extracted and updated, and then propagate into answers. Its remedy is to store the right memories. We study what remains when the stored memory \emph{is} right: the model can still ignore it, misapply it to another entity, or answer from a superseded version. These use-time memory hallucinations are context-conflicting hallucinations caused by the conflict between parametric knowledge and the agent's own memory. They cannot be fixed by better storage alone, which is why \method intervenes in how the model \emph{reads} a memory.

\paragraph{Text memory for agents.}
Most agent memories store natural-language records and retrieve them into the prompt. Examples are retrieval-augmented generation~\citep{lewis2020rag}, operating-system-style paging~\citep{packer2023memgpt}, reflective memory streams~\citep{park2023generative}, forgetting-curve stores~\citep{zhong2024memorybank}, and production systems that extract, consolidate and link memories~\citep{chhikara2025mem0,xu2025amem}. Long-conversation benchmarks such as LoCoMo~\citep{maharana2024locomo} and LongMemEval~\citep{wu2025longmemeval} measure whether the right memory is retrieved and answered from. These systems optimise \emph{what reaches the context}. \method is complementary: it keeps a text store and controls \emph{how the model uses} what reaches the context. On LoCoMo we compare against Mem0 and A-Mem with the same answering model.

\paragraph{Parametric memory: editing, parametric RAG and memory-native models.}
\emph{Knowledge editing} writes facts into weights. Locate-then-edit methods treat MLPs as key--value stores~\citep{geva2021ffn} and insert closed-form updates: ROME~\citep{meng2022rome}, MEMIT~\citep{meng2023memit}, and AlphaEdit~\citep{fang2025alphaedit} with null-space constraints. Memory-based editors keep side memories with scope classifiers or codebooks: SERAC~\citep{mitchell2022serac}, GRACE~\citep{hartvigsen2023grace}, WISE~\citep{wang2024wise}, Larimar~\citep{das2024larimar}. IKE~\citep{zheng2023ike} edits in context, and RECIPE~\citep{chen2024recipe}, DeCK~\citep{bi2024deck} and LightEdit~\citep{jung2026lightedit} combine retrieval with prompts or decoding that suppress the old fact. Sequential editing remains fragile~\citep{lin2024sequential,yao2023editing}.
Our \E follows the ROME lineage; what differs is how \method uses it. The edit's subject key also serves as the address that gates text and steering. Each memory keeps its own patch, loaded only when its address fires, and a patch is retired when its memory is revised.
\emph{Parametric RAG} turns documents into LoRA weights generated by a hypernetwork and combines them with text RAG~\citep{tan2025dyprag,tang2025pkrag}. This resembles our \T+\E, but without per-query arbitration.
\emph{Memory-native models} make memory part of the architecture. MemoryLLM~\citep{wang2024memoryllm} keeps a self-updating memory pool inside the transformer, and Temp-LoRA~\citep{wang2024templora} trains a temporary adapter on the history. MemT-1.0~\citep{memtensor2026memt} is the most recent and most scaled example. It adds native memory blocks to selected layers. A dynamic per-trajectory memory state is written token by token with a gated delta rule and read through a parallel memory attention, and Mixture-of-Storage scales its capacity. The memory modules are trained with the backbone frozen, on 34B tokens of continued pre-training and post-training, by distilling the backbone that sees the full textual history. After writing, the history text is no longer needed in context. MemT-1.0 and \method both bring memory into the model's computation, but they put different things there (\cref{tab:position}). MemT-1.0 internalises the \emph{content} of the history in a trained, fixed-size state; it is evaluated on recall and use of that history, and its training target is the backbone's own full-context behaviour. \method leaves the content in an inspectable text store and writes into the model only a per-memory \emph{use policy}, an address and an adopt/restrain steering pair, which act on a frozen, off-the-shelf model without any pre-training. The two are therefore orthogonal. \method targets the arbitration failures of \cref{fig:conflict}, which full-context imitation does not address by construction. A memory-native model could also expose its memory readout to the same kind of address-gated steering.

\paragraph{Knowledge conflicts and context faithfulness.}
LLMs can over-rely on parametric knowledge and ignore context~\citep{longpre2021entity}. They can also be very receptive to coherent but wrong context~\citep{xie2024chameleon}. Both behaviours are surveyed by \citet{xu2024conflicts}. Existing remedies push the model towards the context, through contrastive decoding (CAD)~\citep{shi2024cad}, attention-head or SAE-feature interventions~\citep{li2025juice,zhao2025spare}, adaptive gating of parametric participation~\citep{li2026spark}, or global steering directions that favour the context, such as ContextFocus~\citep{anand2026contextfocus} and the knowledge-selection direction (KSD) of \citet{lou2026ksd}, a signed diff-of-means direction that transfers across conflict types and models. These methods treat faithfulness as a global property of a prompt. In an agent memory the same note is relevant to some queries and irrelevant to others, so a global push turns parametric override into memory override. \method makes the decision per memory and per query: the address selects between an adopt and a restrain vector. We compare directly against CAD, KSD, ContextFocus and CAST on five models.

\paragraph{Activation steering, and steering as memory.}
Adding vectors to the residual stream steers behaviour: ActAdd~\citep{turner2023actadd}, CAA~\citep{rimsky2024caa}, ITI~\citep{li2023iti}, representation engineering~\citep{zou2023repe}. CAST~\citep{lee2025cast} applies a behaviour vector only when the hidden state is similar to a condition vector. We share the idea of gating, but our gate is a per-memory address taken from the edit key. Our steered behaviour is evidence use, not refusal.
Several works store knowledge \emph{as} activations or representation edits: SAKE~\citep{scialanga2025sake}, REACT~\citep{zhong2025react}, MEGA~\citep{cao2026mega}, RILKE~\citep{liu2025rilke}. Neural Procedural Memory~\citep{zhao2026npm} stores agent skills as retrieved steering vectors and finds them complementary to textual workflow memory. \citet{jumle2026steeringmemory} show that facts barely survive when stored as steering alone (at most 15\% clean use, against 96--98\% in context). This negative result is consistent with our design choice: \method never asks \Sv to carry content (\Sv alone: 1.7\%), only to control how text content is used.
\citet{chavira2026controlled} argue conceptually that memory agents should divide labour among retrieval, steering and weight editing. \method gives one concrete instance of that division, with experiments.

\begin{table}[t]
\centering\small
\caption{\textbf{Where the memory lives and who arbitrates.} ``Content in model'': whether the fact or history itself is written into weights or a trained state. ``Arbitration'': how the system decides, per query, between the memory and parametric knowledge.}
\label{tab:position}
\renewcommand{\arraystretch}{1.15}
\resizebox{\linewidth}{!}{%
\begin{tabular}{llllll}
\toprule
Paradigm & Memory content & Written into the model & Base-model training & Arbitration with parametric knowledge & Per-memory removal \\
\midrule
Text memory / RAG~\citep{lewis2020rag,chhikara2025mem0,xu2025amem} & text & nothing & none & left to the model & delete record \\
Knowledge editing~\citep{meng2022rome,meng2023memit,fang2025alphaedit} & weights & fact (weight delta) & none & edit replaces parametric fact & partly (side memories) \\
Context-faithfulness steering~\citep{anand2026contextfocus,lou2026ksd,shi2024cad} & text & one global direction / decoding rule & none & always favour context & -- \\
Parametric RAG~\citep{tan2025dyprag} & text + generated LoRA & document content & hypernetwork & left to the model & drop adapter \\
MemT-1.0~\citep{memtensor2026memt} & trained native memory state & history content & 34B-token memory training & imitate full-context backbone & no (fixed-size state) \\
\rowcolor{ours}\method (ours) & text (+ local patch) & address + adopt/restrain policy & none (frozen) & per query, address-gated $\vp$/$\vm$ & yes (versioned, unloadable) \\
\bottomrule
\end{tabular}}
\end{table}

\section{Method: TASK Memory}
\label{sec:method}

\subsection{Setting and notation}
An agent runs a frozen decoder-only LM $f_\theta$ with residual stream $h^{(\ell)}_t\in\mathbb R^d$ at layer $\ell$ and token $t$. During deployment it receives \emph{observations}, i.e.\ facts it is allowed to see, and later \emph{queries}. A structured memory $m$ comes from one observation: a subject $s_m$, a relation prompt $p_m$ that contains $s_m$, and a new object $o^*_m$. A free-form memory is a sentence extracted from a dialogue session.
\paragraph{Use-time memory hallucination.} Let $o^{\text{old}}_m$ be the pre-update object, and let $\mathcal N_m$ be the neighbour queries that are not about $m$ but that the no-memory model answers correctly. Given a correct memory $m$ in the store, an answer $y$ to query $x$ is a \emph{memory-ignoring} hallucination (Type I) if $x$ is about $m$ and $y$ asserts $o^{\text{old}}_m$ instead of $o^*_m$; a \emph{memory-misapplying} hallucination (Type II) if $x\in\mathcal N_m$ and $y$ asserts $o^*_m$, attributing the memory to the wrong entity; and a \emph{version-confusion} hallucination (Type III) if $m$ has been revised to $m'$ and $y$ asserts $o^*_m$ instead of $o^*_{m'}$. Avoiding the first two requires \emph{adoption}, answering $o^*_m$ on queries about $m$ even when the weights hold the stale object, and \emph{restraint}, answering as the model would without $m$ on queries that are not about $m$, even when $m$ is in the context. \method assigns these requirements to different components: the steering pair provides adoption and restraint against Types I and II, the address decides which memories are in play at all, and supersession at revision time removes Type III.

\subsection{The memory object}
\paragraph{\T: content.} A JSONL record holds the fact text, scope, version, source, a retrieval key (subject and relation cue) and a \texttt{supersedes} link. \T is the only carrier of content in \method memory.

\paragraph{Address.} Let $k_t\in\mathbb R^{d_\text{ff}}$ be the input of the down-projection of the MLP at layer $\ell_E$ at token $t$. Following the key--value view of MLPs~\citep{geva2021ffn,meng2022rome}, the key of $m$ is
\[
k_m=\textstyle\mathbb E_{c}\big[k_{t_s}(c)\big],
\]
the average of $k_{t_s}$ at the last subject token $t_s$ over a set of writer contexts $c$. These contexts are the relation prompt, mined same-relation templates from the training split, the evidence-in-context form, and generic Wikitext prefixes. We estimate the second moment $C=\mathbb E[kk^\top]$ once on about $7\times10^5$ Wikitext tokens~\citep{merity2017wikitext} and set
\begin{equation}
u_m=C^{-1}k_m,\qquad \hat u_m=\frac{u_m}{k_m^\top u_m},\qquad r_m(x)=\max_t\ \hat u_m^\top k_t(x).
\label{eq:address}
\end{equation}
By construction $r_m=1$ on the memory's own key. Directions that ordinary text uses heavily are down-weighted by $C^{-1}$. The threshold $\tau_m$ is calibrated per memory at write time: it is the midpoint between the minimum response on the memory's own contexts and the maximum response on hard negatives (same relation, other subjects). The gate is $g_m(x)=\mathbb 1[r_m(x)\ge\tau_m]$.
For free-form facts, where the subject rarely appears verbatim in questions, we use a semantic address in the same spirit. The writer asks the model for three questions that the fact answers. The key is the centred, normalised mean-pooled layer-$\ell_S$ state of these questions and the statement. $\tau_m$ is calibrated against other memories' self-questions, and the most similar candidate always passes.

\paragraph{\E: a removable closed-form patch on the same key.} Following ROME~\citep{meng2022rome}, we optimise a residual $\delta_m$ at the output of layer $\ell_E$ with Adam (25 steps, lr 0.5). The objective is the likelihood of $o^*_m$ across the writer contexts, a KL term that keeps ``$\{s_m\}$ is a'' unchanged, weight decay and a norm cap. The rank-one update
\begin{equation}
\Delta W_m=\frac{\delta_m\,u_m^\top}{k_m^\top u_m}\quad\text{satisfies}\quad \Delta W_m k_m=\delta_m .
\label{eq:edit}
\end{equation}
For several memories we solve $\Delta W=R(\lambda I+K^\top C^{-1}K)^{-1}(C^{-1}K)^\top$ with residuals $R$ and keys $K$. Each patch is stored as low-rank factors and applied by a hook as the effective weight $W+BA$. It is bound to the model hash and can be unloaded bit-exactly. In \method, \E plays two roles. Its key \emph{is} the address of \cref{eq:address}, and its weight change provides associative recall that multi-hop questions can chain through (\cref{sec:results}).

\paragraph{\Sv: a paired use policy.} Each memory gets two vectors $\vp_m,\vm_m\in\mathbb R^d$. They are added to the residual stream at the output of layer $\ell_S$, only at the \emph{decision positions}, i.e.\ the last prompt token and all generated tokens, so the encoding of the question itself is untouched:
\begin{equation}
h^{(\ell_S)}_t\leftarrow h^{(\ell_S)}_t+g_m(x)\,\vp_m+\big(1-g_m(x)\big)\,\vm_m .
\label{eq:steer}
\end{equation}
Only the two vectors are trained; $\theta$ is frozen, but gradients still flow through it. The objective has three terms:
\begin{align}
\mathcal L_m=\;&\underbrace{\mathbb E_{c\in\mathcal C^+_m}\big[-\log p_{\theta,\vp_m}(o^*_m\mid \T_m,c)\big]}_{\text{adopt}}
+\underbrace{\mathbb E_{c\in\mathcal C^-_m}\big[\mathrm{KL}\big(p_\theta(\cdot\mid c)\,\|\,p_{\theta,\vm_m}(\cdot\mid \T_m,c)\big)\big]}_{\text{restrain}}
+\lambda\frac{\|\vp_m\|^2+\|\vm_m\|^2}{\mathrm{rms}^2_{\ell_S}} .
\label{eq:sloss}
\end{align}
The adopt term is taken over contexts $\mathcal C^+_m$ that ask about $s_m$ with the note present, where the gate is on and the gradient reaches $\vp_m$. The restrain term is taken over hard negatives $\mathcal C^-_m$ that share the relation but not the subject and contain $m$'s note, which is exactly where text memory is misused; there the gate is off, and $\vm_m$ is trained to reproduce what the model says \emph{without} the note. Vectors are zero-initialised and trained with Adam for 40 steps at lr $0.02\times\mathrm{rms}_{\ell_S}$. We call this version S$_\text{ind}$. A variant, S$_\text{cond}$, trains the vectors on the edited model $\theta+\Delta W_m$ to test whether steering should compensate for the residual errors of the edit (\cref{sec:rq1}); unless stated otherwise \method uses S$_\text{ind}$.

\subsection{Reading: one address, three gates}
Given a query $x$, BM25~\citep{robertson2009bm25} first returns candidates: the top two for structured memories, and the union of the BM25 and semantic top five for dialogue facts. One partial forward pass to layer $\ell_E$ then yields $r_m(x)$ for every candidate. Only memories with $g_m=1$ are shown as text and have their own single-memory patches loaded (summed; ``gated \E''), and their steering is applied by \cref{eq:steer}. A single context manager owns all patch and steering hooks and removes them after generation. Relevance is thus decided once, by the model's own representation, so the three interventions cannot contradict each other: a note is never shown without its patch, and an adopt vector is never applied to an unrelated query.

\paragraph{Read rule for composition.} $\vm$ assumes that a query can be attributed to individual memories. In multi-hop questions the bridge entity does not appear in the question, so $\vm$ would suppress evidence that is needed. For multi-hop composition we therefore apply only $\vp$, and paired steering elsewhere.

\subsection{Revision and supersession}
When a new observation arrives, BM25 retrieves the top two active records, the observation's address response is checked against them, and the agent's own frozen model judges with a fixed prompt whether the observation gives a new value for the same entity and attribute (\textsc{update}) or states a different fact (\textsc{different}). On \textsc{update}, the old record is superseded, and its \E and \Sv leave the active set together; no fact IDs, relation IDs or version numbers are involved. Retiring the old \E is necessary because two versions have almost identical keys, and a joint closed-form solve over both collapses. Retiring the old \Sv proves equally necessary, since otherwise the adopt vectors of both versions fire and each pushes its own object (\cref{sec:rq2}).

\section{Experimental Setup}
\label{sec:setup}

\paragraph{Models.} We use five frozen open models from four families: Qwen3-8B (primary)~\citep{yang2025qwen3}, Llama-3.1-8B-Instruct~\citep{grattafiori2024llama3}, Qwen3-14B, Phi-4 (14B)~\citep{abdin2024phi4} and Gemma-3-27B~\citep{gemma2025gemma3}, all run in BF16 with greedy decoding (non-thinking mode for Qwen3). For each model, $\ell_E$ is chosen on training-split worlds by one fixed rule, edit-only paraphrase accuracy over three candidate layers, which gives layers 6/4/9/7/11 in the order above; $\ell_S$ is set by relative depth (14/13/16/16/25). All other hyperparameters were chosen once on Qwen3-8B development data and reused unchanged for every model.

\paragraph{Benchmarks and protocols.} The protocols vary along two axes: how the memory arrives (provided, retrieved from a library, or accumulated with revisions) and what form the fact takes (structured or free-form). The \emph{provided-evidence} protocol isolates the use of a memory: CounterFact~\citep{meng2022rome} and zsRE~\citep{levy2017zsre} each contribute 300 test worlds with one memory per world, and the note is in the context for both related and unrelated queries. In the \emph{library} protocol, one agent holds all memories of the split (300, and 1k and 3k with training-split distractors) and must retrieve before reading. The \emph{continual-revision} protocol inserts 100 facts and then revises 50 of them. Multi-hop composition is evaluated on MQuAKE-CF-3k~\citep{zhong2023mquake} (200 cases, 600 questions), where only the edited base facts are written. Free-form dialogue memory is evaluated on LoCoMo~\citep{maharana2024locomo}, split by conversation into 2 development and 8 test conversations with 1{,}230 test questions of categories 1--4. Results on LongMemEval with oracle histories~\citep{wu2025longmemeval} and on AppWorld~\citep{trivedi2024appworld} with a local ReAct agent~\citep{yao2023react} are in \cref{app:more}.

\paragraph{Metrics.} \emph{Adoption} is paraphrase and canonical accuracy on queries about the memory. \emph{Restraint} is measured by locality preserved, i.e.\ one minus the damage rate on neighbour queries that the no-memory model answers correctly, and by \emph{misuse}, the rate at which a neighbour answer asserts the memory's object. These quantities map directly onto the hallucination types of \cref{sec:method}. The Type I rate is \emph{reversion}, the fraction of answers that assert the pre-update object, reported on queries whose stale fact is stored in the weights; the Type II rate is misuse; and the Type III rate is the \emph{stale-version rate}, the fraction of answers about a revised fact that assert the superseded value. The \emph{composite} is the harmonic mean of paraphrase accuracy, canonical accuracy and locality preserved, so it is low whenever one of the three is low. Answers are classified by a normalised assertion matcher that ignores negated mentions. On LoCoMo we report LLM-judge accuracy under three judges (Qwen3-8B, Qwen3-14B and GLM-5.3) together with input tokens.

\paragraph{Baselines.} Six context-use methods receive the same note as \method: IKE~\citep{zheng2023ike}, CAD~\citep{shi2024cad}, CAA~\citep{rimsky2024caa}, KSD~\citep{lou2026ksd}, ContextFocus~\citep{anand2026contextfocus} and CAST~\citep{lee2025cast}, with strengths tuned once on Qwen3-8B development data. The parametric carriers are ROME, MEMIT, AlphaEdit and WISE in their original EasyEdit implementations~\citep{wang2024easyedit}. For dialogue we compare with Mem0~\citep{chhikara2025mem0} and A-Mem~\citep{xu2025amem}, both ingesting every turn with the same local model, and with the full conversation in context. Control conditions replace \Sv with norm-matched random vectors, vectors at the wrong layer, another memory's vectors, or an ungated global evidence vector.

\paragraph{Protocol and statistics.} Train/dev/test splits are hashed by subject (or by conversation) and frozen before any model output. Hyperparameters and read rules were selected on train or dev data, the primary test comparisons were registered before their outputs existed, and each test set was run once. Confidence intervals are 95\% paired-bootstrap intervals over worlds (or conversations, or subjects in clustered settings) with 2{,}000 resamples~\citep{efron1994bootstrap}; ``significant'' ($^{*}$) means that the interval excludes 0.

\section{Results}
\label{sec:results}

We organise the results around three questions. RQ1 asks whether, once a correct memory is in context, \method prevents memory-ignoring (Type I) and memory-misapplying (Type II) hallucinations. RQ2 asks whether, when one agent holds many memories that keep changing, the shared address reads the right ones and avoids version confusion (Type III). RQ3 asks whether updated facts can be composed with what the model already knows. \Cref{tab:headline} lists the pre-registered test comparisons on the primary model.

\begin{table}[t]
\centering\small
\caption{\textbf{Pre-registered test comparisons} (Qwen3-8B; one frozen configuration; each test set run once). $\Delta$ in percentage points with 95\% paired-bootstrap CI; bold and $^{*}$: CI excludes 0. The upper bound 0.00 in the last row is exactly zero, so that interval includes 0. Composite = harmonic mean of paraphrase accuracy, canonical accuracy and locality preserved.}
\label{tab:headline}
\resizebox{\linewidth}{!}{%
\begin{tabular}{llllccr}
\toprule
& Benchmark & \method configuration & Baseline & \method & Baseline & $\Delta$ [95\% CI] \\
\midrule
\multirow{3}{*}{RQ1} & CounterFact (300 worlds) & T + S & T & 0.919 & 0.811 & \textbf{+10.80} \ci{7.80}{14.16}\sig \\
& zsRE (300 worlds) & T + S & T & 0.892 & 0.835 & \textbf{+5.65} \ci{1.20}{10.20}\sig \\
& LoCoMo (1{,}230 q.) & retrieved T + S & retrieved T & 0.342 & 0.328 & \textbf{+1.38} \ci{0.21}{2.75}\sig \\
\midrule
\multirow{4}{*}{RQ2} & CounterFact library (300) & address-verified T + gated E + S & BM25 RAG & 0.946 & 0.782 & \textbf{+16.36} \ci{12.99}{20.31}\sig \\
& zsRE library (300) & address-verified T + gated E + S & BM25 RAG & 0.965 & 0.803 & \textbf{+16.23} \ci{11.29}{22.11}\sig \\
& LoCoMo & semantic-address T + S & BM25 RAG & 0.342 & 0.308 & \textbf{+3.41} \ci{0.07}{7.05}\sig \\
& LoCoMo & semantic-address T (text only) & BM25 RAG & 0.328 & 0.308 & +2.03 \ci{-1.21}{5.51} \\
\midrule
RQ3 & MQuAKE-CF-3k (600 q.) & T + E + S($\vp$) & T & 0.765 & 0.690 & \textbf{+7.50} \ci{3.33}{11.83}\sig \\
\midrule
\multirow{2}{*}{Analysis} & CounterFact & T + E + S$_\text{cond}$ & T + E + S$_\text{ind}$ & 0.883 & 0.867 & +1.55 \ci{-1.41}{4.22} \\
& zsRE & T + E + S$_\text{cond}$ & T + E + S$_\text{ind}$ & 0.882 & 0.891 & $-$0.91 \ci{-3.30}{0.00} \\
\bottomrule
\end{tabular}}
\end{table}

\subsection{RQ1: Preventing memory-ignoring and memory-misapplying hallucinations}
\label{sec:rq1}

\begin{table}[t]
\centering\small
\caption{\textbf{Provided evidence} (test, Qwen3-8B). The note is in context for related \emph{and} unrelated queries. Para./Canon.: adoption on paraphrased/canonical queries (\%); Loc.: locality preserved (\%); Misuse: neighbour answers that assert the memory's object (\%, all neighbours); Comp.: composite.}
\label{tab:provided}
\resizebox{\linewidth}{!}{%
\begin{tabular}{lccccc@{\hskip 12pt}ccccc}
\toprule
& \multicolumn{5}{c}{CounterFact (300)} & \multicolumn{5}{c}{zsRE (300)} \\
\cmidrule(lr){2-6}\cmidrule(lr){7-11}
Method & Para. & Canon. & Loc. & Misuse$\downarrow$ & Comp. & Para. & Canon. & Loc. & Misuse$\downarrow$ & Comp. \\
\midrule
No memory & 1.8 & 0.7 & 100.0 & 1.7 & 0.015 & 1.7 & 1.3 & 100.0 & 0.0 & 0.022 \\
T (text memory) & 88.2 & 91.7 & 67.8 & 39.2 & 0.811 & 91.7 & 92.7 & 70.4 & 5.7 & 0.835 \\
\midrule
T + global evidence vector (ungated) & 88.2 & 93.3 & 60.9 & 43.5 & 0.780 & 93.0 & 95.0 & 66.7 & 8.3 & 0.827 \\
T + random vector (norm-matched) & 88.3 & 91.0 & 69.1 & 38.8 & 0.816 & 92.7 & 92.3 & 70.4 & 5.0 & 0.837 \\
T + vector at wrong layer & 88.3 & 91.7 & 68.7 & 39.2 & 0.816 & 91.7 & 92.7 & 68.5 & 5.7 & 0.827 \\
T + another memory's S & 91.5 & 94.7 & 78.3 & 12.8 & 0.875 & 92.3 & 93.7 & 72.2 & 1.0 & 0.849 \\
S only (no text) & 1.7 & 0.7 & 83.9 & 1.5 & 0.014 & 1.0 & 1.3 & 87.0 & 0.0 & 0.017 \\
\rowcolor{ours}\textbf{T + S (\method)} & 96.7 & 99.3 & 81.7 & 7.5 & \textbf{0.919} & 98.7 & 100.0 & 74.1 & 2.3 & \textbf{0.892} \\
\midrule
E only & 79.7 & 95.3 & 94.8 & 2.8 & 0.893 & 72.0 & 93.0 & 100.0 & 0.0 & 0.866 \\
T + E & 96.8 & 99.0 & 50.4 & 52.7 & 0.745 & 97.3 & 99.0 & 68.5 & 5.0 & 0.858 \\
T + E + S$_\text{ind}$ & 97.3 & 99.7 & 70.0 & 16.3 & 0.867 & 98.3 & 100.0 & 74.1 & 2.7 & 0.891 \\
T + E + S$_\text{cond}$ & 97.3 & 99.0 & 73.5 & 10.3 & 0.883 & 98.3 & 100.0 & 72.2 & 1.7 & 0.882 \\
\bottomrule
\end{tabular}}
\end{table}

\paragraph{\Sv improves adoption and restraint at the same time.} On CounterFact (\cref{tab:provided}), \Sv raises paraphrase adoption from 88.2\% to 96.7\%. It also cuts neighbour misuse from 39.2\% to 7.5\% and raises locality from 67.8\% to 81.7\%: +10.8\pp composite. A single direction cannot achieve both: the ungated evidence vector increases misuse to 43.5\% and \emph{lowers} the composite below text memory, while norm-matched random vectors and wrong-layer vectors leave behaviour unchanged. Without the note, \Sv answers 1.7\% of queries, the same as the unmodified model, so its effect lies in how the note is used rather than in any stored content.

\begin{figure}[t]
\centering
\includegraphics[width=\linewidth]{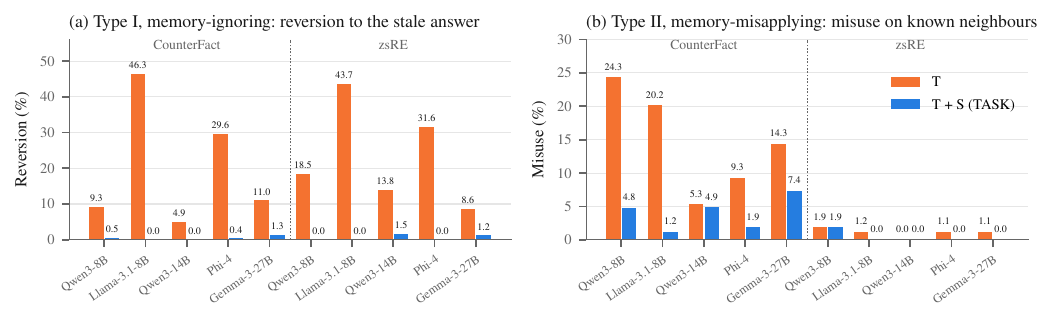}
\caption{\textbf{\method reduces both types of use-time memory hallucination} on five models and two datasets (test). \textbf{(a)} Type I, memory-ignoring: reversion to the stale parametric answer on queries whose stale fact is in the weights. \textbf{(b)} Type II, memory-misapplying: misuse on neighbour questions that the no-memory model answers correctly. This subset is why the values differ from the misuse over all neighbours in \cref{tab:provided} (e.g.\ 24.3\% vs.\ 39.2\% for Qwen3-8B with \T). Bars give values in \%. Per-cell bootstrap CIs are in \cref{app:conflict}.}
\label{fig:resolution}
\end{figure}

\paragraph{Both types of memory hallucination fall at once.} \Cref{fig:resolution} revisits the two hallucination types of \cref{fig:conflict} on all five models. For Type I, reversion to the stale answer on queries whose stale fact is stored in the weights falls from 4.9--46.3\% under \T to 0--1.5\% under \T+\Sv, significantly in all ten model--dataset cells; on Llama-3.1-8B CounterFact, for example, it drops from 46.3\% to 0\% ($-$46.3\pp \ci{-53.5}{-39.1}). The adoption gap between stored and not-stored facts, which \T shows in five cells, disappears, and no cell has a positive significant gap under \T+\Sv. For Type II, misuse on CounterFact neighbours that the model knows falls from 5.3--24.3\% to 1.2--7.4\%, significantly on four of five models; on the fifth, Qwen3-14B, \T misuses only 5.3\% to begin with. That both failures shrink together, which no single global direction can achieve, indicates that \Sv learns an \emph{arbitration} policy rather than a bias towards context.

\paragraph{What \Sv learns: a shared direction behind a per-memory gate.} We decompose every memory's vectors into the mean over all 300 memories (``shared'') and a residual (Appendix \cref{tab:sana}). With each memory's own gate kept, the shared direction alone reaches 0.909, statistically indistinguishable from full \Sv (0.919; full \Sv $-$ shared $=+0.99$\pp \ci{-1.07}{3.00}); the memory-specific residual alone reaches 0.866; and another memory's vectors recover about 60\% of the gain (0.875). Most of what \Sv encodes is therefore a reusable ``adopt / restrain the note'' direction, with a smaller memory-specific part. What makes this direction useful is the \emph{gate}: the same kind of direction applied without a per-memory gate (ContextFocus, KSD) or behind a generic hidden-state condition (CAST) is significantly worse in 9 of 10 model--dataset cells (\cref{sec:general}). The address, not the vector, is where per-memory specificity lives.

\paragraph{Edit-aware steering is not needed.} Training \Sv on the edited model (S$_\text{cond}$) was the original motivation for coupling steering with editing. S$_\text{cond}$ does depend on its edit: unloading \E lowers its dev paraphrase accuracy from 197 to 187 of 200. It also asserts the new object on fewer neighbours (62 vs.\ 98 of 600). But its composite advantage over S$_\text{ind}$ is not significant on CounterFact ($+$1.55\pp) or zsRE ($-$0.91\pp). Across eleven settings it is significantly positive once and significantly negative three times (Appendix \cref{tab:gcond}). We therefore use S$_\text{ind}$. Most of what \Sv learns is decoupled from the particular edit.

\subsection{RQ2: Reading the right memories from a growing library}
\label{sec:rq2}

\begin{table}[t]
\centering\small
\caption{\textbf{One agent, 300 memories} (test; BM25 top-2 candidates) and \textbf{continual revision} (dev; 100 insertions then 50 revisions). E$_\text{all}$: one joint patch over all memories, always on.}
\label{tab:library}
\begin{subtable}[t]{0.58\linewidth}
\centering
\resizebox{\linewidth}{!}{%
\begin{tabular}{lcccccc}
\toprule
& \multicolumn{3}{c}{CounterFact library} & \multicolumn{3}{c}{zsRE library} \\
\cmidrule(lr){2-4}\cmidrule(lr){5-7}
Method & Para. & Loc. & Comp. & Para. & Loc. & Comp. \\
\midrule
BM25 RAG (T) & 83.3 & 64.9 & 0.782 & 85.7 & 68.4 & 0.803 \\
BM25 RAG + S & 80.8 & 78.5 & 0.837 & 91.7 & 70.2 & 0.846 \\
BM25 RAG + E$_\text{all}$ & 93.2 & 51.3 & 0.736 & 91.0 & 70.2 & 0.843 \\
Address-verified T & 82.2 & 96.9 & 0.898 & 89.7 & 94.7 & 0.923 \\
Address-verified T + S & 89.3 & 96.9 & 0.950 & 96.3 & 94.7 & 0.970 \\
Address-verified T + gated E & 89.7 & 94.7 & 0.942 & 94.3 & 94.7 & 0.959 \\
\rowcolor{ours}\textbf{Addr.-verified T + gated E + S} & 90.0 & 95.2 & 0.946 & 95.3 & 94.7 & 0.965 \\
\bottomrule
\end{tabular}}
\end{subtable}\hfill
\begin{subtable}[t]{0.40\linewidth}
\centering
\resizebox{\linewidth}{!}{%
\begin{tabular}{lccc}
\toprule
Method & Current & Stale$\downarrow$ & Loc. \\
\midrule
BM25 RAG, naive append & 68.7 & 50.7 & 45.0 \\
BM25 RAG + supersession & 94.3 & 0.7 & 53.3 \\
RAG + one E over all versions & 1.3 & 3.3 & 0.0 \\
Address-verified T & 93.0 & 0.0 & 93.3 \\
\rowcolor{ours}\textbf{Addr.-verified T + gated E} & 99.0 & 0.0 & 95.0 \\
\quad + S & 98.7 & 0.0 & 95.0 \\
\bottomrule
\end{tabular}}
\end{subtable}
\end{table}

\paragraph{The address prevents memory-misapplying hallucinations in a library.} With 300 memories, BM25 retrieves a distractor note for many unrelated queries; the model then applies it, and locality drops to 64.9\% (\cref{tab:library}, left). Verifying candidates with the in-model address raises locality to 96.9\%. Adding \Sv on top lifts adoption: +16.4\pp composite on CounterFact and +16.2\pp on zsRE over BM25 RAG. Without the address, weight memories do not scale: an always-on joint patch drops locality to 51.3\%, whereas the gated per-memory patches keep it at about 95\%. In this setting \T+\Sv and \T+gated \E+\Sv perform alike (0.950 vs.\ 0.946). On structured single facts the address is what \E contributes; its weight change adds nothing.

\begin{figure}[t]
\centering
\includegraphics[width=\linewidth]{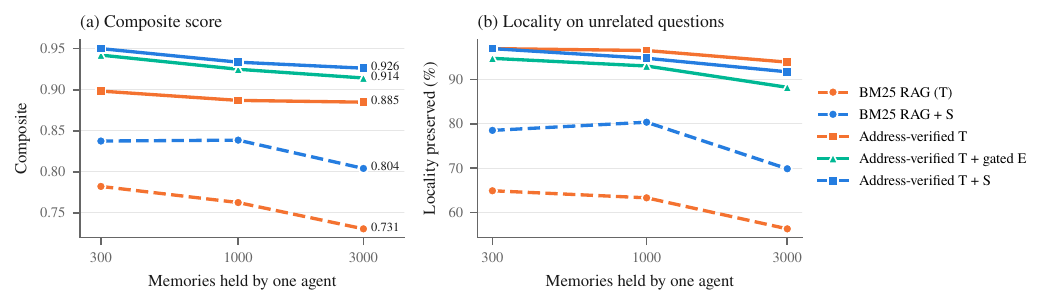}
\caption{\textbf{Scaling the library to 3{,}000 memories} (CounterFact test queries; distractors are training-split memories written by the same writer). Dashed: BM25 retrieval; solid: address-verified reading. Colours follow the carriers as in \cref{fig:conflict,fig:resolution}. At 300 memories, address-verified \T and \T+\Sv have the same locality (96.9\%), so their markers coincide in (b). The full configuration with gated \E (address-verified \T + gated \E + \Sv, \cref{tab:library}) scores 0.946 / 0.928 / 0.916 and lies between the two upper curves in (a); it is omitted for legibility.}
\label{fig:scaling}
\end{figure}

\paragraph{Scaling.} From 300 to 3{,}000 memories (\cref{fig:scaling}), BM25 RAG degrades from 0.782 to 0.731, because locality falls to 56.3\%. Address-verified \T+\Sv stays at 0.926, with locality above 91\%, and the full configuration with gated \E stays at 0.916.

\paragraph{Version-confusion hallucinations: supersession is a precondition for steering.} Naive appending leaves both versions of a fact in the library, and half of the answers assert the old value (\cref{tab:library}, right). Placing both versions in one joint closed-form patch collapses it (current-value accuracy 1.3\%). With address verification and gated \E, current-value accuracy is 99.0\% with no stale answers. On the harder test version of this setting (\cref{app:hard}), all Qwen3-8B methods drop to 33--42\% accuracy on revised facts when versions coexist. On Llama-3.1-8B, \Sv even \emph{amplifies} the conflict ($-$28.7\pp): the adopt vectors of both versions fire, and each pushes its own object. Revision must therefore retire the old \Sv together with the old \T and \E. Our automatic supersession, which combines the address with the agent's own judge and uses no fact IDs, recovers this. It improves over not superseding by 48.7\pp on Qwen3-8B and 54.7\pp on Llama-3.1-8B (both significant), matches oracle links exactly on Llama-3.1-8B, and falls 10.7\pp short of them on Qwen3-8B, whose judge misses 14\% of true updates. In terms of Type III hallucination, the stale-version rate of address-verified \T+\Sv on revised facts falls from 54.7\% to 10.0\% on Qwen3-8B and from 48.0\% to 0\% on Llama-3.1-8B (\cref{tab:h2}).

\subsection{RQ3: Composing updated facts}
\label{sec:rq3}

\begin{table}[t]
\centering\small
\caption{\textbf{Composition and the parametric carrier.} \emph{Left:} MQuAKE-CF-3k test (200 cases, 600 questions; only edited base facts are written). \emph{Right:} our closed-form \E against the original EasyEdit implementations on zsRE test (300 edits).}
\label{tab:compose}
\begin{subtable}[t]{0.5\linewidth}
\centering
\resizebox{\linewidth}{!}{%
\begin{tabular}{lccc}
\toprule
Method & Multi-hop & Case & Old-answer$\downarrow$ \\
\midrule
No memory & 2.8 & 4.0 & 27.5 \\
T & 69.0 & 79.5 & 6.5 \\
T + S (paired) & 62.2 & 75.0 & 9.2 \\
T + S ($\vp$ only) & 69.3 & 80.5 & 6.3 \\
E only & 5.5 & 8.0 & 14.3 \\
T + E & 76.8 & 85.5 & 3.2 \\
\rowcolor{ours}\textbf{T + E + S ($\vp$ only)} & 76.5 & 85.0 & 3.0 \\
\bottomrule
\end{tabular}}
\end{subtable}\hfill
\begin{subtable}[t]{0.47\linewidth}
\centering
\resizebox{\linewidth}{!}{%
\begin{tabular}{lccc}
\toprule
Edit method & E only & T + E & T + E + S \\
\midrule
EasyEdit ROME & 0.756 & 0.836 & 0.860 \\
EasyEdit MEMIT & 0.739 & 0.852 & 0.876 \\
EasyEdit AlphaEdit & 0.720 & 0.855 & 0.856 \\
\rowcolor{ours}Ours (closed-form, subject key) & 0.866 & 0.858 & 0.891 \\
\bottomrule
\end{tabular}}
\end{subtable}
\end{table}

Multi-hop questions require chaining an updated fact through the model's own knowledge, and here the weight carrier matters (\cref{tab:compose}, left). Text memory reaches 69.0\% multi-hop accuracy, and adding \E raises it to 76.8\%; \method (\T+\E+\Sv$^{+}$) reaches 76.5\%, +7.5\pp \ci{3.3}{11.8}. The gain is due to \E, not \Sv. Reversion to the old answer halves (6.5\%$\to$3.0\%). Paired steering \emph{hurts} here (62.2\%): the bridge entity is not in the question, so $\vm$ suppresses notes that are needed. This motivates the $\vp$-only read rule for composition. As a carrier, our subject-keyed closed-form patch is stronger than the original ROME, MEMIT and AlphaEdit implementations under the same requests (\cref{tab:compose}, right). \Sv also improves every one of these editors when stacked on top.

\subsection{Generality across models and against baselines}
\label{sec:general}

\begin{table}[t]
\centering\small
\caption{\textbf{Same frozen protocol on five models} (test). \emph{Top:} gains of \method. $\Delta_1$: T+S $-$ T; $\Delta_2$: address-verified T + gated E + S $-$ BM25 RAG (300-memory library); $\Delta_3$: T+E+S$^{+}$ $-$ T (MQuAKE). \emph{Bottom:} T+S (\method) minus each context-use baseline with the same note (composite, pp). Bold and $^{*}$: 95\% CI excludes 0, in either direction (full CIs in \cref{app:more}). n/a: not run for Gemma-3-27B, as fixed in advance for compute reasons.}
\label{tab:general}
\resizebox{\linewidth}{!}{%
\begin{tabular}{lccccccc}
\toprule
Model & CF: T & CF: T+S & CF $\Delta_1$ & CF lib.\ $\Delta_2$ & zsRE $\Delta_1$ & zsRE lib.\ $\Delta_2$ & MQuAKE $\Delta_3$ \\
\midrule
Qwen3-8B & 0.811 & 0.919 & \textbf{+10.80}\sig & \textbf{+16.36}\sig & \textbf{+5.65}\sig & \textbf{+16.23}\sig & \textbf{+7.50}\sig \\
Llama-3.1-8B & 0.518 & 0.948 & \textbf{+43.02}\sig & \textbf{+49.37}\sig & \textbf{+28.04}\sig & \textbf{+41.40}\sig & \textbf{+33.83}\sig \\
Qwen3-14B & 0.923 & 0.945 & \textbf{+2.19}\sig & \textbf{+6.05}\sig & \textbf{+7.90}\sig & \textbf{+11.60}\sig & +1.17 \\
Phi-4 (14B) & 0.756 & 0.956 & \textbf{+19.97}\sig & \textbf{+12.99}\sig & \textbf{+8.20}\sig & \textbf{+20.12}\sig & \textbf{+36.00}\sig \\
Gemma-3-27B & 0.849 & 0.888 & \textbf{+3.92}\sig & \textbf{+17.65}\sig & $-$4.61 & n/a & n/a \\
\midrule
\midrule
Model $\cdot$ dataset & $-$IKE & $-$CAD & $-$CAA & $-$KSD & $-$ContextFocus & $-$CAST & \\
\midrule
Qwen3-8B $\cdot$ CF & \textbf{+34.67}\sig & \textbf{+23.44}\sig & \textbf{+8.89}\sig & \textbf{+11.51}\sig & \textbf{+10.60}\sig & \textbf{+10.37}\sig & \\
Qwen3-8B $\cdot$ zsRE & +0.36 & \textbf{+15.16}\sig & \textbf{+9.96}\sig & \textbf{+6.18}\sig & \textbf{+11.37}\sig & \textbf{+5.84}\sig & \\
Llama-3.1-8B $\cdot$ CF & \textbf{+39.37}\sig & \textbf{+55.06}\sig & \textbf{+38.37}\sig & \textbf{+41.37}\sig & \textbf{+41.90}\sig & \textbf{+36.90}\sig & \\
Llama-3.1-8B $\cdot$ zsRE & \textbf{+27.06}\sig & \textbf{+53.21}\sig & \textbf{+43.02}\sig & \textbf{+27.56}\sig & \textbf{+24.27}\sig & \textbf{+26.29}\sig & \\
Qwen3-14B $\cdot$ CF & \textbf{+2.73}\sig & \textbf{+7.20}\sig & \textbf{+2.41}\sig & \textbf{+2.66}\sig & \textbf{+2.19}\sig & \textbf{+2.09}\sig & \\
Qwen3-14B $\cdot$ zsRE & $-$4.06 & \textbf{+11.32}\sig & \textbf{+12.04}\sig & \textbf{+6.75}\sig & \textbf{+8.04}\sig & \textbf{+7.12}\sig & \\
Phi-4 $\cdot$ CF & \textbf{+5.91}\sig & \textbf{+21.48}\sig & \textbf{+24.41}\sig & \textbf{+19.21}\sig & \textbf{+19.28}\sig & \textbf{+21.14}\sig & \\
Phi-4 $\cdot$ zsRE & $-$1.31 & \textbf{+16.29}\sig & \textbf{+17.45}\sig & \textbf{+8.64}\sig & \textbf{+15.83}\sig & \textbf{+8.43}\sig & \\
Gemma-3-27B $\cdot$ CF & \textbf{+26.37}\sig & \textbf{+13.86}\sig & \textbf{+3.52}\sig & \textbf{+3.53}\sig & \textbf{+3.92}\sig & \textbf{+3.68}\sig & \\
Gemma-3-27B $\cdot$ zsRE & \textbf{$-$9.71}\sig & $-$0.69 & $-$2.35 & $-$4.61 & $-$2.60 & $-$4.25 & \\
\bottomrule
\end{tabular}}
\end{table}

\paragraph{Across models.} The same frozen protocol holds on all five models (\cref{tab:general}, top): 21 of 23 cross-model comparisons are significantly positive. The gain is largest where the model resists its context most. Llama-3.1-8B adopts text memory poorly (CF composite 0.518) and gains +43.0\pp. Qwen3-14B already follows context well (0.923) and gains +2.2\pp. This matches the conflict analysis: models with larger parametric override (\cref{fig:conflict}a) leave more room for arbitration.

\paragraph{Against context-use baselines.} With the same note in context, \method significantly beats the three steering methods closest to it (KSD, ContextFocus, CAST) in 9 of 10 model--dataset cells, and CAA in 9 of 10 (\cref{tab:general}, bottom). The exception is Gemma-3-27B on zsRE. There \Sv raises paraphrase accuracy (0.907$\to$0.967) but increases locality damage, and no difference to the steering baselines is significant. IKE, which adds demonstrations and is especially strong on zsRE's question format, ties with \method or wins on 14B+ models for zsRE. On CounterFact, IKE's demonstrations cause heavy misuse, and \method leads by 2.7--39.4\pp.

\subsection{Free-form dialogue memory}
\label{sec:dialogue}

\begin{table}[t]
\centering\small
\caption{\textbf{LoCoMo test} (8 conversations, 1{,}230 questions; same answering model, Qwen3-8B). Accuracy under three judges and mean input tokens. Bold: best among methods with at most about 300 input tokens. Paired differences use a conversation-level bootstrap.}
\label{tab:locomo}
\resizebox{0.86\linewidth}{!}{%
\begin{tabular}{lcccr}
\toprule
Method & Qwen3-8B judge & Qwen3-14B judge & GLM-5.3 judge & Input tokens \\
\midrule
BM25 top-5 facts (T) & 30.8 & 28.5 & 33.6 & 299 \\
Semantic-address T & 32.8 & 30.3 & 34.6 & 298 \\
\rowcolor{ours}\textbf{Semantic-address T + S (\method)} & \textbf{34.2} & \textbf{31.0} & \textbf{39.3} & 298 \\
Mem0, top-5 & 17.9 & 16.1 & 19.5 & 257 \\
Mem0, top-30 & 26.1 & 24.2 & 30.4 & 1{,}226 \\
A-Mem, token-matched to ours & 28.6 & 27.1 & 30.7 & 294 \\
A-Mem, top-5 & 39.4 & 37.5 & 42.8 & 1{,}248 \\
A-Mem, top-30 & 51.5 & 48.0 & 55.9 & 6{,}845 \\
Full conversation in context (reference) & 60.3 & 57.5 & 64.2 & 19{,}488 \\
\midrule
\method $-$ semantic-address T & +1.38\sig & +0.65 & +4.63\sig & \\
\method $-$ Mem0 top-5 & +16.34\sig & +14.88\sig & +19.76\sig & \\
\method $-$ A-Mem token-matched & +5.61\sig & +3.90\sig & +8.62\sig & \\
\bottomrule
\end{tabular}}
\end{table}

On LoCoMo (\cref{tab:locomo}), facts are free-form and the subject key does not apply, so \method uses the semantic address and \T+\Sv, without \E. At an equal input budget of about 300 tokens, \method improves over BM25 retrieval, over the same retrieval without \Sv (significant under two of three judges), over Mem0, and over token-matched A-Mem (significant under all three judges). The absolute level stays modest: A-Mem with 4--23$\times$ more tokens and the full conversation (65$\times$) remain clearly more accurate. On LongMemEval the literal subject key gates out relevant facts and \E/\Sv give no gain (\cref{app:more}), which is what motivated the semantic address.

\section{Discussion and Limitations}
\label{sec:discussion}

\paragraph{What each carrier contributes.} The evidence assigns each carrier a distinct role. Text carries the content: without it, neither \Sv nor single-fact accuracy survives. Steering carries the use policy and is the most consistent source of gains, resolving both directions of the conflict on all five models. The address taken from \E's key controls which memories are read; it turns retrieval into precise reading for libraries of up to 3{,}000 memories and is what allows weight patches to scale. \E's weight change matters where facts must be composed (MQuAKE) but adds nothing to \T+\Sv on single facts: on CounterFact, \T+\E+\Sv scores below \T+\Sv on four models and ties on Gemma-3-27B (0.887 vs.\ 0.888), and in libraries the two tie. \method memory is therefore best understood as a division of labour, in which each carrier is used for what it does well.

\paragraph{Memory hallucination beyond storage.} Our results suggest that memory approaches should be judged on \emph{use} as well as on \emph{storage}: whether a stored item is adopted where it applies and ignored where it does not. Storage-stage benchmarks such as HaluMem~\citep{chen2025halumem} and our use-time analysis cover complementary halves of memory hallucination. A memory system can store perfectly and still hallucinate at read time. Writing content into weights (editing) or into a trained state (MemT-1.0) changes where the memory lives. It does not by itself arbitrate between the memory and the model's prior knowledge. Combined with retrieval, editing even amplifies memory override. An address-gated use policy addresses this arbitration directly; because it acts on a frozen model through one hook and one partial forward pass, it can be combined with either kind of storage.

\paragraph{Limitations.} The gains of \method memory shrink when a model already follows its context well (Qwen3-14B CounterFact: +2.2\pp), and on Gemma-3-27B zsRE \T+\Sv is not better than \T ($-$4.61\pp \ci{-10.25}{0.01}), while IKE is significantly better. On dialogue memory the gains are small and \method stays far below full-context reading and token-rich A-Mem; the literal subject key fails on free-form questions (LongMemEval), and the semantic address fixes this only in part. Revision depends on supersession, without which \Sv can amplify version conflicts, and the automatic judge is imperfect: it misses 14\% of true updates on Qwen3-8B, and in libraries with many facts per subject the Llama-3.1-8B judge wrongly supersedes 34\% of memories, while about 10\% of sibling memories remain confused under every carrier. Writing a memory costs about 25\,s of steering training and 4\,s of editing on one GPU, and our libraries reach 3{,}000 memories rather than millions; the decomposition in \cref{sec:rq1} suggests that a shared direction behind per-memory gates could remove most of this cost. Finally, CounterFact, zsRE and MQuAKE consist of synthetic counterfactual updates, our answer classifier and LLM judges are not the official scorers, and on AppWorld the local 8B agent almost never completes tasks, with \Sv improving only the partial test-pass rate (+2.2\pp, not significant).

\paragraph{Broader impact.} Making agents adopt stored notes more reliably also makes them adopt \emph{wrong} notes more reliably. \method should sit behind a writer that verifies sources. Its per-memory, versioned and removable design makes such auditing and deletion straightforward.

\section{Conclusion}
\label{sec:conclusion}
Text memory tells a model \emph{what} a fact is, but not \emph{whether} the fact should override what the model already believes. We showed that this gap produces use-time memory hallucinations, even when the stored memory is correct. The model ignores the memory in favour of stale weights, misapplies it to other entities, or answers from a superseded version. The first two come from the conflict between parametric knowledge and external memory, which runs in both directions on every model we tested. \method memory addresses it with a per-memory use policy behind an in-model address: paired adopt/restrain steering vectors, gated by the key of a removable knowledge edit, with content kept in text. Under a pre-registered protocol on five frozen models, this policy resolves both conflict directions and makes memory libraries precise up to 3{,}000 memories; the edit carrier adds compositional recall. By making each memory more than text, a \textbf{T}ext note combined with \textbf{A}ctivation \textbf{S}teering and \textbf{K}nowledge editing, each in the role it is good at, we can reduce memory hallucinations without retraining the model. We hope these results encourage treating memory \emph{use}, and not only memory \emph{storage}, as a first-class target for reducing hallucination in LLM agents.

\bibliographystyle{plainnat}
\bibliography{references}

@inproceedings{lewis2020rag,
  title={Retrieval-Augmented Generation for Knowledge-Intensive {NLP} Tasks},
  author={Lewis, Patrick and Perez, Ethan and Piktus, Aleksandra and Petroni, Fabio and Karpukhin, Vladimir and Goyal, Naman and K{\"u}ttler, Heinrich and Lewis, Mike and Yih, Wen-tau and Rockt{\"a}schel, Tim and Riedel, Sebastian and Kiela, Douwe},
  booktitle={Advances in Neural Information Processing Systems},
  year={2020}
}

@article{packer2023memgpt,
  title={{MemGPT}: Towards {LLMs} as Operating Systems},
  author={Packer, Charles and Wooders, Sarah and Lin, Kevin and Fang, Vivian and Patil, Shishir G. and Stoica, Ion and Gonzalez, Joseph E.},
  journal={arXiv preprint arXiv:2310.08560},
  year={2023}
}

@inproceedings{park2023generative,
  title={Generative Agents: Interactive Simulacra of Human Behavior},
  author={Park, Joon Sung and O'Brien, Joseph and Cai, Carrie Jun and Morris, Meredith Ringel and Liang, Percy and Bernstein, Michael S.},
  booktitle={Proceedings of the 36th Annual ACM Symposium on User Interface Software and Technology},
  year={2023}
}

@inproceedings{zhong2024memorybank,
  title={{MemoryBank}: Enhancing Large Language Models with Long-Term Memory},
  author={Zhong, Wanjun and Guo, Lianghong and Gao, Qiqi and Ye, He and Wang, Yanlin},
  booktitle={Proceedings of the AAAI Conference on Artificial Intelligence},
  year={2024}
}

@article{chhikara2025mem0,
  title={{Mem0}: Building Production-Ready {AI} Agents with Scalable Long-Term Memory},
  author={Chhikara, Prateek and Khant, Dev and Aryan, Saket and Singh, Taranjeet and Yadav, Deshraj},
  journal={arXiv preprint arXiv:2504.19413},
  year={2025}
}

@article{xu2025amem,
  title={{A-MEM}: Agentic Memory for {LLM} Agents},
  author={Xu, Wujiang and Liang, Zujie and Mei, Kai and Gao, Hang and Tan, Juntao and Zhang, Yongfeng},
  journal={arXiv preprint arXiv:2502.12110},
  year={2025}
}

@inproceedings{maharana2024locomo,
  title={Evaluating Very Long-Term Conversational Memory of {LLM} Agents},
  author={Maharana, Adyasha and Lee, Dong-Ho and Tulyakov, Sergey and Bansal, Mohit and Barbieri, Francesco and Fang, Yuwei},
  booktitle={Proceedings of the 62nd Annual Meeting of the Association for Computational Linguistics},
  year={2024}
}

@inproceedings{wu2025longmemeval,
  title={{LongMemEval}: Benchmarking Chat Assistants on Long-Term Interactive Memory},
  author={Wu, Di and Wang, Hongwei and Yu, Wenhao and Zhang, Yuwei and Chang, Kai-Wei and Yu, Dong},
  booktitle={International Conference on Learning Representations},
  year={2025}
}

@inproceedings{trivedi2024appworld,
  title={{AppWorld}: A Controllable World of Apps and People for Benchmarking Interactive Coding Agents},
  author={Trivedi, Harsh and Khot, Tushar and Hartmann, Mareike and Manku, Ruskin and Dong, Vinty and Li, Edward and Gupta, Shashank and Sabharwal, Ashish and Balasubramanian, Niranjan},
  booktitle={Proceedings of the 62nd Annual Meeting of the Association for Computational Linguistics},
  year={2024}
}

@inproceedings{yao2023react,
  title={{ReAct}: Synergizing Reasoning and Acting in Language Models},
  author={Yao, Shunyu and Zhao, Jeffrey and Yu, Dian and Du, Nan and Shafran, Izhak and Narasimhan, Karthik and Cao, Yuan},
  booktitle={International Conference on Learning Representations},
  year={2023}
}

@article{robertson2009bm25,
  title={The Probabilistic Relevance Framework: {BM25} and Beyond},
  author={Robertson, Stephen and Zaragoza, Hugo},
  journal={Foundations and Trends in Information Retrieval},
  volume={3},
  number={4},
  pages={333--389},
  year={2009}
}

@inproceedings{geva2021ffn,
  title={Transformer Feed-Forward Layers Are Key-Value Memories},
  author={Geva, Mor and Schuster, Roei and Berant, Jonathan and Levy, Omer},
  booktitle={Proceedings of the 2021 Conference on Empirical Methods in Natural Language Processing},
  year={2021}
}

@inproceedings{meng2022rome,
  title={Locating and Editing Factual Associations in {GPT}},
  author={Meng, Kevin and Bau, David and Andonian, Alex and Belinkov, Yonatan},
  booktitle={Advances in Neural Information Processing Systems},
  year={2022}
}

@inproceedings{meng2023memit,
  title={Mass-Editing Memory in a Transformer},
  author={Meng, Kevin and Sharma, Arnab Sen and Andonian, Alex and Belinkov, Yonatan and Bau, David},
  booktitle={International Conference on Learning Representations},
  year={2023}
}

@inproceedings{fang2025alphaedit,
  title={{AlphaEdit}: Null-Space Constrained Knowledge Editing for Language Models},
  author={Fang, Junfeng and Jiang, Houcheng and Wang, Kun and Ma, Yunshan and Wang, Xiang and He, Xiangnan and Chua, Tat-Seng},
  booktitle={International Conference on Learning Representations},
  year={2025}
}

@inproceedings{wang2024wise,
  title={{WISE}: Rethinking the Knowledge Memory for Lifelong Model Editing of Large Language Models},
  author={Wang, Peng and Li, Zexi and Zhang, Ningyu and Xu, Ziwen and Yao, Yunzhi and Jiang, Yong and Xie, Pengjun and Huang, Fei and Chen, Huajun},
  booktitle={Advances in Neural Information Processing Systems},
  year={2024}
}

@inproceedings{hartvigsen2023grace,
  title={Aging with {GRACE}: Lifelong Model Editing with Discrete Key-Value Adaptors},
  author={Hartvigsen, Thomas and Sankaranarayanan, Swami and Palangi, Hamid and Kim, Yoon and Ghassemi, Marzyeh},
  booktitle={Advances in Neural Information Processing Systems},
  year={2023}
}

@inproceedings{mitchell2022serac,
  title={Memory-Based Model Editing at Scale},
  author={Mitchell, Eric and Lin, Charles and Bosselut, Antoine and Manning, Christopher D. and Finn, Chelsea},
  booktitle={International Conference on Machine Learning},
  year={2022}
}

@inproceedings{zheng2023ike,
  title={Can We Edit Factual Knowledge by In-Context Learning?},
  author={Zheng, Ce and Li, Lei and Dong, Qingxiu and Fan, Yuxuan and Wu, Zhiyong and Xu, Jingjing and Chang, Baobao},
  booktitle={Proceedings of the 2023 Conference on Empirical Methods in Natural Language Processing},
  year={2023}
}

@inproceedings{yao2023editing,
  title={Editing Large Language Models: Problems, Methods, and Opportunities},
  author={Yao, Yunzhi and Wang, Peng and Tian, Bozhong and Cheng, Siyuan and Li, Zhoubo and Deng, Shumin and Chen, Huajun and Zhang, Ningyu},
  booktitle={Proceedings of the 2023 Conference on Empirical Methods in Natural Language Processing},
  year={2023}
}

@inproceedings{zhong2023mquake,
  title={{MQuAKE}: Assessing Knowledge Editing in Language Models via Multi-Hop Questions},
  author={Zhong, Zexuan and Wu, Zhengxuan and Manning, Christopher D. and Potts, Christopher and Chen, Danqi},
  booktitle={Proceedings of the 2023 Conference on Empirical Methods in Natural Language Processing},
  year={2023}
}

@inproceedings{levy2017zsre,
  title={Zero-Shot Relation Extraction via Reading Comprehension},
  author={Levy, Omer and Seo, Minjoon and Choi, Eunsol and Zettlemoyer, Luke},
  booktitle={Proceedings of the 21st Conference on Computational Natural Language Learning},
  year={2017}
}

@inproceedings{wang2024easyedit,
  title={{EasyEdit}: An Easy-to-use Knowledge Editing Framework for Large Language Models},
  author={Wang, Peng and Zhang, Ningyu and Tian, Bozhong and Xi, Zekun and Yao, Yunzhi and Xu, Ziwen and Wang, Mengru and Mao, Shengyu and Wang, Xiaohan and Cheng, Siyuan and Liu, Kangwei and Ni, Yuansheng and Zheng, Guozhou and Chen, Huajun},
  booktitle={Proceedings of the 62nd Annual Meeting of the Association for Computational Linguistics: System Demonstrations},
  year={2024}
}

@inproceedings{das2024larimar,
  title={Larimar: Large Language Models with Episodic Memory Control},
  author={Das, Payel and Chaudhury, Subhajit and Nelson, Elliot and Melnyk, Igor and Swaminathan, Sarathkrishna and Dai, Sihui and Lozano, Aur{\'e}lie and Kollias, Georgios and Chenthamarakshan, Vijil and Navr{\'a}til, Ji{\v{r}}{\'\i} and Dan, Soham and Chen, Pin-Yu},
  booktitle={International Conference on Machine Learning},
  year={2024}
}

@inproceedings{wang2024memoryllm,
  title={{MEMORYLLM}: Towards Self-Updatable Large Language Models},
  author={Wang, Yu and Gao, Yifan and Chen, Xiusi and Jiang, Haoming and Li, Shiyang and Yang, Jingfeng and Yin, Qingyu and Li, Zheng and Li, Xian and Yin, Bing and Shang, Jingbo and McAuley, Julian},
  booktitle={International Conference on Machine Learning},
  year={2024}
}

@article{wang2024templora,
  title={With Greater Text Comes Greater Necessity: Inference-Time Training Helps Long Text Generation},
  author={Wang, Y. and Ma, D. and Cai, D.},
  journal={arXiv preprint arXiv:2401.11504},
  year={2024}
}

@article{tan2025dyprag,
  title={Dynamic Parametric Retrieval Augmented Generation for Test-time Knowledge Enhancement},
  author={Tan, Yuqiao and He, Shizhu and Liao, Huanxuan and Zhao, Jun and Liu, Kang},
  journal={arXiv preprint arXiv:2503.23895},
  year={2025}
}

@article{tang2025pkrag,
  title={Understanding Parametric Knowledge Injection in Retrieval-Augmented Generation},
  author={Tang, Minghao and Ni, Shiyu and others},
  journal={arXiv preprint arXiv:2510.12668},
  year={2025}
}

@article{memtensor2026memt,
  title={{MemT-1.0}: Towards Memory-Native Evolution},
  author={{MemTensor Team}},
  journal={arXiv preprint arXiv:2610.26825},
  year={2026}
}

@inproceedings{chen2024recipe,
  title={Lifelong Knowledge Editing for {LLMs} with Retrieval-Augmented Continuous Prompt Learning},
  author={Chen, Qizhou and Zhang, Taolin and He, Xiaofeng and Li, Dongyang and Wang, Chengyu and Huang, Longtao and Xue, Hui},
  booktitle={Proceedings of the 2024 Conference on Empirical Methods in Natural Language Processing},
  year={2024}
}

@article{bi2024deck,
  title={Decoding by Contrasting Knowledge: Enhancing {LLMs}' Confidence on Edited Facts},
  author={Bi, Baolong and Wang, Yiwei and Liu, Shenghua and Ji, Pengliang and Mei, Lingrui and Cheng, Xueqi},
  journal={arXiv preprint arXiv:2405.11613},
  year={2024}
}

@inproceedings{jung2026lightedit,
  title={Towards Scalable Lifelong Knowledge Editing with Selective Knowledge Suppression},
  author={Jung, Dahyun and Lee, Jaewook and Lim, Heuiseok},
  booktitle={Proceedings of the 64th Annual Meeting of the Association for Computational Linguistics},
  year={2026}
}

@article{lin2024sequential,
  title={Navigating the Dual Facets: A Comprehensive Evaluation of Sequential Memory Editing in Large Language Models},
  author={Lin, Zihao and Beigi, Mohammad and Li, Hongxuan and Zhou, Yufan and Zhang, Yuxiang and Wang, Qifan and Yin, Wenpeng and Huang, Lifu},
  journal={arXiv preprint arXiv:2402.11122},
  year={2024}
}

@inproceedings{scialanga2025sake,
  title={{SAKE}: Steering Activations for Knowledge Editing},
  author={Scialanga, Marco and Laugel, Thibault and Grari, Vincent and Detyniecki, Marcin},
  booktitle={Proceedings of the 63rd Annual Meeting of the Association for Computational Linguistics},
  year={2025}
}

@article{zhong2025react,
  title={{REACT}: Representation Extraction And Controllable Tuning to Overcome Overfitting in {LLM} Knowledge Editing},
  author={Zhong, Haitian and Liu, Yuhuan and Xu, Ziyang and Liu, Guofan and Liu, Qiang and Wu, Shu and Zhao, Zhe and Wang, Liang and Tan, Tieniu},
  journal={arXiv preprint arXiv:2505.18933},
  year={2025}
}

@article{liu2025rilke,
  title={Representation Interventions Enable Lifelong Knowledge Memory Control in {LLMs}},
  author={Liu, Xuyuan and Chen, Shengyu and Dong, Xinshuai and Liu, Yanchi and Zhao, Xujiang and Wang, Haoyu and Yan, Yujun and Chen, Haifeng and Chen, Zhengzhang},
  journal={arXiv preprint arXiv:2511.20892},
  year={2025}
}

@article{cao2026mega,
  title={The Anatomy of an Edit: Mechanism-Guided Activation Steering for Knowledge Editing},
  author={Cao, Yuan and Wang, Mingyang and Sch{\"u}tze, Hinrich},
  journal={arXiv preprint arXiv:2603.20795},
  year={2026}
}

@article{chavira2026controlled,
  title={Controlled Knowledge Updating in Memory-Enabled {AI} Agents: Beyond Model Editing and Recall},
  author={Chavira Ju{\'a}rez, Gabriel and Gonzalez Bravo, Eder Jahir and Rivera Garc{\'\i}a, Guadalupe Esmeralda and Arcos Espinosa, Javier A. and Nava-D{\'\i}az, Salvador W.},
  journal={Frontiers in Big Data},
  volume={9},
  pages={1938137},
  year={2026}
}

@inproceedings{longpre2021entity,
  title={Entity-Based Knowledge Conflicts in Question Answering},
  author={Longpre, Shayne and Perisetla, Kartik and Chen, Anthony and Ramesh, Nikhil and DuBois, Chris and Singh, Sameer},
  booktitle={Proceedings of the 2021 Conference on Empirical Methods in Natural Language Processing},
  year={2021}
}

@inproceedings{xie2024chameleon,
  title={Adaptive Chameleon or Stubborn Sloth: Revealing the Behavior of Large Language Models in Knowledge Conflicts},
  author={Xie, Jian and Zhang, Kai and Chen, Jiangjie and Lou, Renze and Su, Yu},
  booktitle={International Conference on Learning Representations},
  year={2024}
}

@inproceedings{xu2024conflicts,
  title={Knowledge Conflicts for {LLMs}: A Survey},
  author={Xu, Rongwu and Qi, Zehan and Guo, Zhijiang and Wang, Cunxiang and Wang, Hongru and Zhang, Yue and Xu, Wei},
  booktitle={Proceedings of the 2024 Conference on Empirical Methods in Natural Language Processing},
  year={2024}
}

@inproceedings{shi2024cad,
  title={Trusting Your Evidence: Hallucinate Less with Context-aware Decoding},
  author={Shi, Weijia and Han, Xiaochuang and Lewis, Mike and Tsvetkov, Yulia and Zettlemoyer, Luke and Yih, Wen-tau},
  booktitle={Proceedings of the 2024 Conference of the North American Chapter of the Association for Computational Linguistics},
  year={2024}
}

@article{anand2026contextfocus,
  title={{ContextFocus}: Activation Steering for Contextual Faithfulness in Large Language Models},
  author={Anand, Nikhil and Somasundaram, Shwetha and Phukan, Anirudh and Saxena, Apoorv and Mukherjee, Koyel},
  journal={arXiv preprint arXiv:2601.04131},
  year={2026}
}

@article{lou2026ksd,
  title={Which Knowledge Should {LLMs} Trust? Uncovering a Generalizable Steering Direction Across Knowledge Conflicts},
  author={Lou, Yibin and Hu, Yi and Chen, Jianhui and Wang, Chenpeng and Zhang, Muhan and Pan, Liangming},
  journal={arXiv preprint arXiv:2610.20067},
  year={2026}
}

@inproceedings{zhao2025spare,
  title={Steering Knowledge Selection Behaviours in {LLMs} via {SAE}-Based Representation Engineering},
  author={Zhao, Yu and Devoto, Alessio and Hong, Giwon and Du, Xiaotang and Gema, Aryo Pradipta and Wang, Hongru and He, Xuanli and Wong, Kam-Fai and Minervini, Pasquale},
  booktitle={Proceedings of the 2025 Conference of the Nations of the Americas Chapter of the Association for Computational Linguistics},
  year={2025}
}

@article{li2025juice,
  title={Taming Knowledge Conflicts in Language Models},
  author={Li, Gaotang and Chen, Yuzhong and Tong, Hanghang},
  journal={arXiv preprint arXiv:2503.10996},
  year={2025}
}

@article{li2026spark,
  title={{SPARK}: Adaptive Parametric Knowledge Participation for Faithful Retrieval-Augmented Generation},
  author={Li, Ruochang and Huang, Pengcheng and Liu, Zhenghao and Yan, Yukun and Xie, Huiyuan and Gu, Yu and Yu, Ge and Sun, Maosong},
  journal={arXiv preprint arXiv:2606.27786},
  year={2026}
}

@article{turner2023actadd,
  title={Steering Language Models With Activation Engineering},
  author={Turner, Alexander Matt and Thiergart, Lisa and Leech, Gavin and Udell, David and Vazquez, Juan J. and Mini, Ulisse and MacDiarmid, Monte},
  journal={arXiv preprint arXiv:2308.10248},
  year={2023}
}

@inproceedings{rimsky2024caa,
  title={Steering {Llama} 2 via Contrastive Activation Addition},
  author={Rimsky, Nina and Gabrieli, Nick and Schulz, Julian and Tong, Meg and Hubinger, Evan and Turner, Alexander Matt},
  booktitle={Proceedings of the 62nd Annual Meeting of the Association for Computational Linguistics},
  year={2024}
}

@inproceedings{li2023iti,
  title={Inference-Time Intervention: Eliciting Truthful Answers from a Language Model},
  author={Li, Kenneth and Patel, Oam and Vi{\'e}gas, Fernanda and Pfister, Hanspeter and Wattenberg, Martin},
  booktitle={Advances in Neural Information Processing Systems},
  year={2023}
}

@article{zou2023repe,
  title={Representation Engineering: A Top-Down Approach to {AI} Transparency},
  author={Zou, Andy and Phan, Long and Chen, Sarah and Campbell, James and Guo, Phillip and Ren, Richard and Pan, Alexander and Yin, Xuwang and Mazeika, Mantas and Dombrowski, Ann-Kathrin and others},
  journal={arXiv preprint arXiv:2310.01405},
  year={2023}
}

@inproceedings{lee2025cast,
  title={Programming Refusal with Conditional Activation Steering},
  author={Lee, Bruce W. and Padhi, Inkit and Natesan Ramamurthy, Karthikeyan and Miehling, Erik and Dognin, Pierre and Nagireddy, Manish and Dhurandhar, Amit},
  booktitle={International Conference on Learning Representations},
  year={2025}
}

@article{zhao2026npm,
  title={Neural Procedural Memory: Empowering {LLM} Agents with Implicit Activation Steering},
  author={Zhao, Chengfeng and Tan, Yuqiao and He, Shizhu and Wang, Yequan and Zhao, Jun and Liu, Kang},
  journal={arXiv preprint arXiv:2606.29824},
  year={2026}
}

@article{jumle2026steeringmemory,
  title={What Survives When Activation Steering Is Used as Memory},
  author={Jumle, Vedant},
  journal={arXiv preprint arXiv:2610.14958},
  year={2026}
}

@article{yang2025qwen3,
  title={{Qwen3} Technical Report},
  author={{Qwen Team}},
  journal={arXiv preprint arXiv:2505.09388},
  year={2025}
}

@article{grattafiori2024llama3,
  title={The {Llama} 3 Herd of Models},
  author={Grattafiori, Aaron and others},
  journal={arXiv preprint arXiv:2407.21783},
  year={2024}
}

@article{abdin2024phi4,
  title={{Phi-4} Technical Report},
  author={Abdin, Marah and others},
  journal={arXiv preprint arXiv:2412.08905},
  year={2024}
}

@article{gemma2025gemma3,
  title={{Gemma} 3 Technical Report},
  author={{Gemma Team}},
  journal={arXiv preprint arXiv:2503.19786},
  year={2025}
}

@inproceedings{merity2017wikitext,
  title={Pointer Sentinel Mixture Models},
  author={Merity, Stephen and Xiong, Caiming and Bradbury, James and Socher, Richard},
  booktitle={International Conference on Learning Representations},
  year={2017}
}

@book{efron1994bootstrap,
  title={An Introduction to the Bootstrap},
  author={Efron, Bradley and Tibshirani, Robert J.},
  publisher={CRC Press},
  year={1994}
}

@article{chen2025halumem,
  title={{HaluMem}: Evaluating Hallucinations in Memory Systems of Agents},
  author={Chen, Ding and Niu, Simin and Li, Zhiyu and others},
  journal={arXiv preprint arXiv:2511.03506},
  year={2025}
}

@article{huang2025survey,
  title={A Survey on Hallucination in Large Language Models: Principles, Taxonomy, Challenges, and Open Questions},
  author={Huang, Lei and Yu, Weijiang and Ma, Weitao and Zhong, Weihong and Feng, Zhangyin and Wang, Haotian and Chen, Qianglong and Peng, Weihua and Feng, Xiaocheng and Qin, Bing and Liu, Ting},
  journal={ACM Transactions on Information Systems},
  volume={43},
  number={2},
  year={2025}
}

@article{zhang2023siren,
  title={Siren's Song in the {AI} Ocean: A Survey on Hallucination in Large Language Models},
  author={Zhang, Yue and Li, Yafu and Cui, Leyang and Cai, Deng and Liu, Lemao and Fu, Tingchen and Huang, Xinting and Zhao, Enbo and Zhang, Yu and Chen, Yulong and Wang, Longyue and Luu, Anh Tuan and Bi, Wei and Shi, Freda and Shi, Shuming},
  journal={arXiv preprint arXiv:2309.01219},
  year={2023}
}

@article{ji2023survey,
  title={Survey of Hallucination in Natural Language Generation},
  author={Ji, Ziwei and Lee, Nayeon and Frieske, Rita and Yu, Tiezheng and Su, Dan and Xu, Yan and Ishii, Etsuko and Bang, Ye Jin and Madotto, Andrea and Fung, Pascale},
  journal={ACM Computing Surveys},
  volume={55},
  number={12},
  pages={1--38},
  year={2023}
}

\clearpage
\appendix
\section{Conflict analysis details}
\label{app:conflict}

\paragraph{Procedure.} The analysis uses only the frozen test predictions of the provided-evidence runs; no model is called again.
\begin{itemize}
\item A paraphrase query counts as \emph{stale fact stored} if the no-memory model's answer to that same query asserts the pre-update object.
\item Neighbour queries count as \emph{base-correct} if the no-memory model answers them correctly.
\item \emph{Gap}: adoption of the memory's object under \T on not-stored queries minus on stored queries.
\item \emph{Reversion}: an answer that asserts the pre-update object.
\item \emph{Misuse}: an answer that asserts the memory's object on a base-correct neighbour.
\end{itemize}
Intervals come from a paired bootstrap over worlds (2{,}000 resamples, seed 17). \Cref{tab:conflict_cells,tab:conflict_rev} are generated by \texttt{scripts/paper\_conflict\_analysis.py}. zsRE neighbours are unrelated questions, so misuse is near zero there for every method.

\begin{table}[h]
\centering\small
\caption{Per-cell conflict statistics (test; \%, unless marked pp).}
\label{tab:conflict_cells}
\resizebox{\linewidth}{!}{
\begin{tabular}{llrcccccccc}
\toprule
 & & & \multicolumn{3}{c}{Parametric override (paraphrase queries)} & \multicolumn{5}{c}{Memory override (base-correct neighbours)} \\
\cmidrule(lr){4-6}\cmidrule(lr){7-11}
Data & Model & $n_\text{stored}$ & T gap (pp) & Rev.\ T & Rev.\ T+S & Mis.\ base & Mis.\ T & Mis.\ T+E & Mis.\ T+S & T+S $-$ T (pp) \\
\midrule
CF & Qwen3-8B & 205/600 & -1.7\,{\scriptsize[-7.3,\,4.4]} & 9.3 & 0.5 & 0.9 & 24.3 & 40.4 & 4.8 & -19.6\,{\scriptsize[-25.8,\,-13.9]}$^{*}$ \\
CF & Llama-3.1-8B & 231/600 & +16.4\,{\scriptsize[8.1,\,25.0]}$^{*}$ & 46.3 & 0.0 & 0.0 & 20.2 & 49.0 & 1.2 & -18.9\,{\scriptsize[-24.9,\,-13.5]}$^{*}$ \\
CF & Qwen3-14B & 223/600 & +0.9\,{\scriptsize[-2.4,\,4.6]} & 4.9 & 0.0 & 0.4 & 5.3 & 11.8 & 4.9 & -0.4\,{\scriptsize[-3.5,\,2.8]} \\
CF & Phi-4 & 257/600 & +14.8\,{\scriptsize[7.2,\,22.4]}$^{*}$ & 29.6 & 0.4 & 0.7 & 9.3 & 19.0 & 1.9 & -7.5\,{\scriptsize[-11.3,\,-4.2]}$^{*}$ \\
CF & Gemma-3-27B & 236/600 & +0.7\,{\scriptsize[-4.8,\,6.6]} & 11.0 & 1.3 & 0.8 & 14.3 & 10.7 & 7.4 & -7.0\,{\scriptsize[-11.4,\,-2.6]}$^{*}$ \\
\midrule
zsRE & Qwen3-8B & 54/300 & +16.9\,{\scriptsize[6.0,\,28.7]}$^{*}$ & 18.5 & 0.0 & 0.0 & 1.9 & 1.9 & 1.9 & +0.0\,{\scriptsize[0.0,\,0.0]} \\
zsRE & Llama-3.1-8B & 87/300 & +28.0\,{\scriptsize[15.7,\,39.8]}$^{*}$ & 43.7 & 0.0 & 0.0 & 1.2 & 7.0 & 0.0 & -1.2\,{\scriptsize[-3.9,\,0.0]} \\
zsRE & Qwen3-14B & 65/300 & +9.1\,{\scriptsize[-0.4,\,19.2]} & 13.8 & 1.5 & 0.0 & 0.0 & 0.0 & 0.0 & +0.0\,{\scriptsize[0.0,\,0.0]} \\
zsRE & Phi-4 & 79/300 & +28.6\,{\scriptsize[16.9,\,40.8]}$^{*}$ & 31.6 & 0.0 & 0.0 & 1.1 & 0.0 & 0.0 & -1.1\,{\scriptsize[-4.0,\,0.0]} \\
zsRE & Gemma-3-27B & 81/300 & +0.7\,{\scriptsize[-6.7,\,7.9]} & 8.6 & 1.2 & 0.0 & 1.1 & 1.1 & 0.0 & -1.1\,{\scriptsize[-4.2,\,0.0]} \\
\bottomrule
\end{tabular}
}
\end{table}

\begin{table}[h]
\centering\small
\caption{Change in reversion to the stale parametric answer from \T to \T+\Sv (stored-fact queries).}
\label{tab:conflict_rev}
\begin{tabular}{llc}
\toprule
Data & Model & Rev.\ T+S $-$ T \\
\midrule
CF & Qwen3-8B & -8.8\,{\scriptsize[-13.8,\,-4.3]}$^{*}$ \\
CF & Llama-3.1-8B & -46.3\,{\scriptsize[-53.5,\,-39.1]}$^{*}$ \\
CF & Qwen3-14B & -4.9\,{\scriptsize[-8.2,\,-2.2]}$^{*}$ \\
CF & Phi-4 & -29.2\,{\scriptsize[-35.7,\,-23.0]}$^{*}$ \\
CF & Gemma-3-27B & -9.7\,{\scriptsize[-14.9,\,-5.1]}$^{*}$ \\
zsRE & Qwen3-8B & -18.5\,{\scriptsize[-29.2,\,-8.5]}$^{*}$ \\
zsRE & Llama-3.1-8B & -43.7\,{\scriptsize[-53.9,\,-33.0]}$^{*}$ \\
zsRE & Qwen3-14B & -12.3\,{\scriptsize[-21.0,\,-4.6]}$^{*}$ \\
zsRE & Phi-4 & -31.6\,{\scriptsize[-42.7,\,-21.8]}$^{*}$ \\
zsRE & Gemma-3-27B & -7.4\,{\scriptsize[-13.6,\,-2.4]}$^{*}$ \\
\bottomrule
\end{tabular}

\end{table}

\section{Implementation details}
\label{app:impl}

\paragraph{Writer.} The writer sees only the update observation (subject, relation prompt and new object for CounterFact/zsRE; the edited base facts for MQuAKE; the dialogue sessions for LoCoMo and LongMemEval). It produces:
\begin{itemize}
\item \T records;
\item four relation templates mined from the training split;
\item hard negatives: same relation, other training-split subjects;
\item generic Wikitext prefixes.
\end{itemize}
Labels enter only after predictions are frozen. Three writer seeds, which resample templates, negatives and prefixes, give consistent \Sv gains on development data (\cref{tab:robust}).

\paragraph{\E.} Value optimisation runs 25 Adam steps at lr 0.5 with early stopping at NLL 0.05, plus KL and norm constraints. The covariance comes from about $7\times10^5$ Wikitext tokens at the edit layer. Patches are stored as low-rank factors.
The edit layer was scanned on training-split smoke worlds. For Qwen3-8B, edit-only paraphrase accuracy was 12/16 at L4, 15/16 at L6 and 12/16 at L8. The other models used the same rule over three candidate layers.
Budget controls (more steps, higher rank) gave no gain on dev. An AlphaEdit-style null-space projection gave identical edits.

\paragraph{\Sv.} 40 Adam steps at lr $0.02\times\mathrm{rms}_{\ell_S}$, zero initialisation, plus a norm penalty. A larger rate (0.05) hurt locality on dev. A residual-weighted variant (fixed a priori) gave identical results.

\paragraph{Intervention checks.} Every intervention changes real activations or effective weights; none is a prompt substitution. In FP32, generation with KV cache under interventions matches full-prefix recomputation (maximum logit difference $8\times10^{-5}$), unloading restores weights bit-exactly, and base parameters receive no gradient during \Sv training. 28 unit tests on a miniature Qwen3 with the real tokenizer cover the gradients of \Sv and \E, causal masking of decision positions, loading and unloading, and hook cleanup.

\paragraph{Cost.} \Cref{tab:cost} lists the measured costs (Qwen3-8B, one A6000 GPU).

\begin{table}[h]
\centering\small
\caption{Write-time and read-time cost per memory (Qwen3-8B, BF16, one GPU).}
\label{tab:cost}
\begin{tabular}{lccc}
\toprule
Carrier / setting & Memories & Seconds / memory & Forward passes / memory \\
\midrule
\E (closed-form patch), CounterFact & 300 & 4.0 & 18.8 \\
\Sv$_\text{ind}$ (paired vectors), CounterFact & 300 & 25.0 & 86 \\
\Sv$_\text{cond}$, CounterFact & 300 & 25.0 & 86 \\
\Sv (semantic address), LoCoMo & 1{,}749 & 25.0 & 86 \\
\bottomrule
\end{tabular}
\end{table}

\section{Hard scenarios and automatic supersession}
\label{app:hard}

\paragraph{Version conflict (H2).} 100 memories are inserted, then 50 of them are revised. We report accuracy on revised facts at the last checkpoint. ``Oracle'' supersession uses the dataset's fact IDs, i.e.\ perfect supersession links. ``Auto'' uses the address together with the agent's own frozen model as judge, with a fixed prompt.

\begin{table}[h]
\centering\small
\caption{Revised facts in the version-conflict test (150 revised facts per model). Accuracy columns give the rate of answers with the current value; stale columns give the rate of answers with the superseded value (Type III). For reference, BM25 RAG without supersession has a stale rate of 0.533 (Qwen3-8B) and 0.140 (Llama-3.1-8B).}
\label{tab:h2}
\resizebox{\linewidth}{!}{%
\begin{tabular}{lcccccc}
\toprule
Model $\cdot$ method & No sup. & No sup.: stale$\downarrow$ & Oracle & Auto & Auto: stale$\downarrow$ & Auto: unrevised \\
\midrule
Qwen3-8B $\cdot$ address-verified T & 0.333 & 0.533 & 0.860 & 0.793 & 0.087 & 0.893 \\
\rowcolor{ours}Qwen3-8B $\cdot$ address-verified T + S & 0.340 & 0.547 & 0.933 & 0.827 & 0.100 & 0.940 \\
Qwen3-8B $\cdot$ address-verified T + gated E + S & 0.420 & 0.473 & 0.940 & 0.893 & 0.053 & 0.953 \\
Llama-3.1-8B $\cdot$ address-verified T & 0.700 & 0.133 & 0.593 & 0.593 & 0.000 & 0.453 \\
\rowcolor{ours}Llama-3.1-8B $\cdot$ address-verified T + S & 0.413 & 0.480 & 0.960 & 0.960 & 0.000 & 0.967 \\
Llama-3.1-8B $\cdot$ address-verified T + gated E + S & 0.380 & 0.500 & 0.953 & 0.953 & 0.000 & 0.967 \\
\bottomrule
\end{tabular}}
\end{table}

\begin{itemize}
\item \textbf{Auto vs.\ none.} Auto supersession improves \T+\Sv over no supersession by +48.67\pp \ci{37.33}{59.33} on Qwen3-8B and +54.67\pp \ci{43.33}{65.33} on Llama-3.1-8B.
\item \textbf{Link quality.} Qwen3-8B's links have precision 1.00 and recall 0.86. Llama-3.1-8B's links are perfect.
\item \textbf{Cost of the address.} The address reduces judge calls from 1.98 to 0.35 per write. Without the address, Llama-3.1-8B's link precision drops to 0.49.
\item \textbf{Without the judge.} Address-only supersession wrongly retires 47\% of memories in a same-subject library.
\end{itemize}

\paragraph{Same-subject library (H1).} 154 subjects with 323 memories; several relations of one subject share a library. Bootstrap is clustered by subject.

\begin{table}[h]
\centering\small
\caption{Same-subject library (accuracy differences, pp). Bold and $^{*}$: 95\% CI excludes 0, in either direction.}
\label{tab:h1}
\begin{tabular}{lcc}
\toprule
Comparison & Qwen3-8B & Llama-3.1-8B \\
\midrule
Addr.\ T+S $-$ addr.\ T & \textbf{+7.02} \ci{4.84}{9.38}\sig & \textbf{+27.55} \ci{23.81}{31.41}\sig \\
Addr.\ T+gated E $-$ addr.\ T & \textbf{+7.22} \ci{4.90}{9.71}\sig & \textbf{+19.61} \ci{15.48}{23.74}\sig \\
Addr.\ T $-$ BM25 T (text only) & \textbf{$-$1.86} \ci{-3.44}{-0.31}\sig & \textbf{$-$4.95} \ci{-7.34}{-2.48}\sig \\
Addr.\ T+gated E+S $-$ addr.\ T+S & +0.10 \ci{-1.12}{1.24} & $-$1.44 \ci{-2.97}{0.10} \\
S$_\text{cond}$ $-$ S$_\text{ind}$ & +0.93 \ci{-0.10}{1.93} & $-$0.93 \ci{-2.63}{0.83} \\
Ungated joint E (accuracy) & 0.015 & 0.013 \\
\bottomrule
\end{tabular}
\end{table}

Sibling memories with the same subject are the hardest case for a subject address. With text only, address verification is slightly worse than BM25. \Sv and \E still give large gains, and gated patches do not collapse the way an ungated joint patch does. About 10\% of answers confuse sibling memories under every carrier.

\section{Additional results}
\label{app:more}

\begin{table}[h]
\centering\small
\caption{Shared/memory-specific decomposition of \Sv (CounterFact test, Qwen3-8B; exploratory). Shared = mean of all 300 memories' $\vp$ (and $\vm$); residual = own vector minus shared. Every variant keeps the memory's own address gate.}
\label{tab:sana}
\begin{tabular}{lcccc}
\toprule
Variant & Para. & Canon. & Loc. & Comp.\ (full S $-$ variant, pp) \\
\midrule
T & 88.2 & 91.7 & 67.8 & 0.811 (+10.80\sig) \\
T + full S & 96.7 & 99.3 & 81.7 & 0.919 \\
T + shared direction only & 93.3 & 97.0 & 83.5 & 0.909 (+0.99 \ci{-1.07}{3.00}) \\
T + memory-specific residual only & 95.0 & 99.0 & 71.3 & 0.866 (+5.32\sig) \\
\bottomrule
\end{tabular}
\\[3pt]{\footnotesize Vector statistics: mean pairwise cosine 0.19 ($\vp$) / 0.19 ($\vm$); cosine to the shared direction 0.42 / 0.44; mean norm 4.7 / 13.0; mean $\cos(\vp_m,\vm_m)=-0.12$.}
\end{table}

\begin{table}[h]
\centering\small
\caption{Edit-aware steering: S$_\text{cond}$ $-$ S$_\text{ind}$ under the same edit (positive favours S$_\text{cond}$). Bold and $^{*}$: 95\% CI excludes 0, in either direction. An upper bound printed as 0.00 is exactly zero, so that interval includes 0.}
\label{tab:gcond}
\begin{tabular}{llcc}
\toprule
Setting & Split & Metric & $\Delta$ (pp) [95\% CI] \\
\midrule
CounterFact, Qwen3-8B & test & composite & +1.55 \ci{-1.41}{4.22} \\
CounterFact, Qwen3-8B, writer seeds 17 / 29 / 41 & dev & composite & +0.84 / $-$1.05 / \textbf{+2.76}\sig \\
CounterFact, Mistral-7B & dev & composite & +1.16 \ci{-3.56}{6.20} \\
CounterFact library, address-verified + gated E & test & composite & +0.01 \ci{-0.51}{0.52} \\
CounterFact library, BM25 + E$_\text{all}$ & test & composite & \textbf{$-$5.18} \ci{-8.26}{-2.33}\sig \\
MQuAKE, paired vectors & test & multi-hop acc. & \textbf{$-$2.67} \ci{-4.50}{-0.67}\sig \\
zsRE, Qwen3-8B & test & composite & $-$0.91 \ci{-3.30}{0.00} \\
zsRE library, address-verified + gated E & test & composite & \textbf{$-$0.45} \ci{-0.95}{-0.11}\sig \\
CounterFact, Llama-3.1-8B / Qwen3-14B / Phi-4 & test & composite & \textbf{+2.87}\sig / +1.30 / +0.11 \\
\bottomrule
\end{tabular}
\end{table}

\begin{table}[h]
\centering\small
\caption{Robustness on development data (100 worlds, composite).}
\label{tab:robust}
\begin{tabular}{lcccc}
\toprule
Method & Qwen3-8B seed 17 & seed 29 & seed 41 & Mistral-7B \\
\midrule
T & 0.825 & 0.825 & 0.825 & 0.639 \\
T + ungated global vector & 0.762 & 0.762 & 0.762 & 0.719 \\
\rowcolor{ours}T + S$_\text{ind}$ & 0.897 & 0.881 & 0.898 & 0.933 \\
T + E & 0.701 & 0.730 & 0.717 & 0.765 \\
T + E + S$_\text{ind}$ & 0.830 & 0.811 & 0.849 & 0.856 \\
E only & 0.890 & 0.891 & 0.898 & 0.898 \\
\bottomrule
\end{tabular}
\\[3pt]{\footnotesize T+S$_\text{ind}$ $-$ T: +7.19, +5.60, +7.33\sig\ (seeds), +29.35\sig\ (Mistral-7B, \ci{22.48}{36.43}).}
\end{table}

\paragraph{Standard editing metrics.} On the frozen test memories, CounterFact ES/PS/NS are computed by comparing the mean-token NLL of the new and the true object, and zsRE is scored by teacher-forced token accuracy. Score is the harmonic mean.
\begin{itemize}
\item \textbf{CounterFact.} \T+\Sv improves the Score over \T on all five models: +16.5 (Qwen3-8B), +33.9 (Llama-3.1-8B), +6.5 (Qwen3-14B), +19.0 (Phi-4), +10.9\pp (Gemma-3-27B), all significant.
\item \textbf{zsRE.} The Score improves on four models (+2.3 to +10.5\pp) and drops on Gemma-3-27B ($-$5.1\pp), consistent with the generation results.
\end{itemize}

\paragraph{WISE.} With hyperparameters mapped from the published configuration, WISE edits only 6 of 200 dev requests in FP32 (BF16 also fails). This is a tuning failure, not an upper bound for WISE, so we exclude it from the main comparison.

\paragraph{LongMemEval (oracle histories).} Same writer, 128 held-out knowledge-update and single-session-user questions.
\begin{itemize}
\item All extracted facts: 99/128, using 533 tokens on average.
\item Full oracle sessions: 109/128, using 4{,}723 tokens.
\item BM25 top-5 facts: 96/128.
\item Literal-key verification: only 39/128, because questions rarely contain the subject verbatim (68/128 questions have any key-verified memory). \E and \Sv add nothing on top of it.
\item Paired \Sv applied to all facts: $-$25\pp. Many simultaneous $\vm$ vectors suppress evidence.
\end{itemize}
The other four question types (352 questions, text methods only) show the same ordering. This negative result motivated the semantic address used on LoCoMo.

\paragraph{AppWorld (dev, 57 tasks).} The agent is a local Qwen3-8B ReAct agent with at most 12 steps. Procedural notes are mined only from train-split solutions. Task success is about 0 for all methods. The mean test-pass rate is 0.252 without memory, 0.259 with \T, and 0.281 with \T+\Sv ($+$2.21\pp \ci{-0.35}{5.41}, not significant).

\paragraph{Full confidence intervals for the steering baselines.} T+S minus KSD, ContextFocus and CAST, with 95\% CIs. The CIs for IKE, CAD and CAA are in our released tables.
\begin{itemize}
\item \textbf{Qwen3-8B.} CF: +11.51 \ci{8.35}{15.22}, +10.60 \ci{7.64}{13.93}, +10.37 \ci{7.35}{13.65}. zsRE: +6.18 \ci{1.47}{11.26}, +11.37 \ci{5.14}{18.77}, +5.84 \ci{1.38}{10.35}.
\item \textbf{Gemma-3-27B, zsRE.} $-$4.61 \ci{-10.25}{0.01}, $-$2.60 \ci{-8.29}{2.40}, $-$4.25 \ci{-9.56}{0.31}.
\end{itemize}
Selected strengths on Qwen3-8B dev:
\begin{itemize}
\item CF: KSD 0.125, ContextFocus 0.125, CAST 4.
\item zsRE: KSD 0.125, ContextFocus 2, CAST 1.
\end{itemize}
CAST's condition vectors reach held-out F1 of 0.956 (CF) and 0.961 (zsRE).

\section{Data boundary and reproducibility}
\label{app:repro}
\begin{itemize}
\item \textbf{Splits.} Splits are hashed by subject (CounterFact, zsRE), by first edited subject (MQuAKE) or by conversation (LoCoMo). Manifests were frozen before any output.
\item \textbf{Writer inputs.} Writers never read test questions, answers, multi-hop targets or test-time scores.
\item \textbf{Pre-registration.} Each test set was run once. Primary comparisons, decision rules and their dates were registered in a decision log before the corresponding outputs existed.
\item \textbf{Denominators.} Failures and runtime errors stay in the denominators. Infrastructure interruptions were resumed, never rerun because of their results.
\item \textbf{Regeneration.} All tables and the data figures are regenerated from raw predictions and metrics files by scripts.
\item \textbf{External API.} Only the GLM-5.3 judge, used to re-judge LoCoMo answers, called an external API.
\end{itemize}

\end{document}